\documentclass[journal,twoside]{IEEEtran}
\usepackage{amsfonts}
\usepackage{cite}
\usepackage[cmex10]{amsmath}
\usepackage{color}
\usepackage{flushend}
\usepackage{url}
\usepackage{array}

\usepackage{mdwmath}
\usepackage{mdwtab}
\usepackage{eqparbox}
\usepackage{tabularx}
\usepackage{booktabs} 

\usepackage{rotating}
\usepackage{afterpage}
\usepackage{placeins}
\usepackage{bbding}
\usepackage{amsmath}   
\usepackage{CJK}    
\usepackage[colorlinks]{hyperref}
\usepackage{lineno}
\usepackage{subcaption}
\usepackage{graphicx}    
\usepackage{multirow}
\usepackage{amssymb}
\usepackage{pifont}

\usepackage{algorithm}
\usepackage{algpseudocode}
\usepackage{bm}

\usepackage{color}\definecolor{MyDarkGreen}{rgb}{0.8,0.91,0.81}\definecolor{yellow}{rgb}{0.99,0.99,0.70}\definecolor{white}{rgb}{1.0,1.0,1.0}\definecolor{black}{rgb}{0.00,0.00,0.00}
\begin{document}

\title{DREB-Net: Dual-stream Restoration Embedding Blur-feature Fusion Network for High-mobility UAV Object Detection}

\author{Qingpeng Li,
	Yuxin Zhang,
    Leyuan Fang,
    Yuhan Kang,
    Shutao Li,
	and~Xiao Xiang Zhu
\thanks{This work was supported in part by the National Natural Science Foundation of China under Grant 62201209, and in part by the Hejian Talents Project Fund under Grant 2023RC3113 of Hunan Province, China.~\emph{(Corresponding author: Leyuan Fang.)}
	
	Qingpeng Li and Yuxin Zhang are with the School of Robotics, Hunan University, 410082 Changsha, China (e-mail: liqingpeng@hnu.edu.cn; zhangyuxin@hnu.edu.cn).
	
	Leyuan Fang, Yuhan Kang and Shutao Li are with the College of Electrical and Information Engineering, Hunan University, 410082 Changsha, China (e-mail: fangleyuan@gmail.com; kyh433@hnu.edu.cn; shutao\_li@hnu.edu.cn).
	
    Xiao Xiang Zhu is with the Department of Data Science in Earth Observation, Technical University Munich, 80333 Munich, Germany (e-mail: xiaoxiang.zhu@tum.de).
	}}
	\markboth{}
	{LI \MakeLowercase{\textit{et al.}}: manuscript to IEEE Transactions on Geoscience and Remote Sensing}
	\maketitle

\begin{abstract}
	Object detection algorithms are pivotal components of unmanned aerial vehicle (UAV) imaging systems, extensively employed in complex fields. However, images captured by high-mobility UAVs often suffer from motion blur cases, which significantly impedes the performance of advanced object detection algorithms. To address these challenges, we propose an innovative object detection algorithm specifically designed for blurry images, named \textbf{DREB-Net} (Dual-stream Restoration Embedding Blur-feature Fusion Network). 
	First, DREB-Net addresses the particularities of blurry image object detection problem by incorporating a Blurry image Restoration Auxiliary Branch (\textbf{BRAB}) during the training phase.
	Second, it fuses the extracted shallow features via Multi-level Attention-Guided Feature Fusion (\textbf{MAGFF}) module, to extract richer features. 
	Here, the MAGFF module comprises local attention modules and global attention modules, which assign different weights to the branches. 
	Then, during the inference phase, the deep feature extraction of the BRAB can be removed to reduce computational complexity and improve detection speed. 
	In loss function, a combined loss of MSE and SSIM is added to the BRAB to restore blurry images. Finally, DREB-Net introduces Fast Fourier Transform in the early stages of feature extraction, via a Learnable Frequency domain Amplitude Modulation Module (\textbf{LFAMM}), to adjust feature amplitude and enhance feature processing capability. 	
	Experimental results indicate that DREB-Net can still effectively perform object detection tasks under motion blur in captured images, showcasing excellent performance and broad application prospects. Our source code will be available at~\url{https://github.com/EEIC-Lab/DREB-Net.git}.
\end{abstract}


\section{Introduction}
\label{sec:introduction}

\IEEEPARstart
{O}{BJECT} detection tasks have long been at the core of research in the field of computer vision. With the rapid development of deep learning technologies, an increasing number of object detection algorithms have been proposed and widely applied across various fields, such as autonomous driving~\cite{feng2020deep}, traffic monitoring~\cite{ghahremannezhad2023object}, disaster relief~\cite{li2018r}, and even consumer electronics~\cite{chen2020mnasfpn}. These applications significantly enhance the system usability and intelligence levels, bringing considerable convenience to people's daily lives.

\begin{figure}[tb!]
	\centering
	\includegraphics[width=88.8mm]{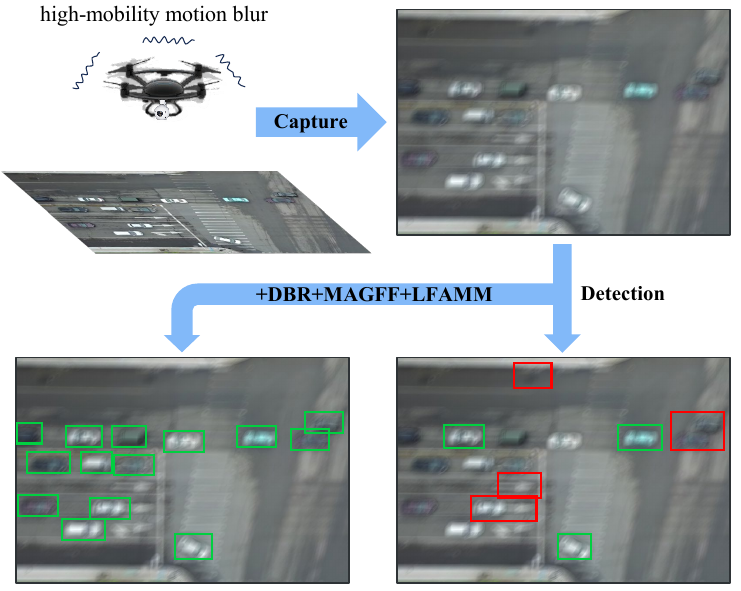}
	\caption{Schematic diagram of the task solved in this paper. Existing object detection algorithms often perform poorly on blurry drone images. Through special design, the object detection algorithm can perform well on high-mobility UAVs.}
	\label{fig1}
\end{figure}

In recent years, with the increasing maturity of cutting-edge technologies such as automatic control~\cite{edgar2000automatic} and image processing~\cite{wei2021fine}, more and more industries have begun to utilize unmanned aerial vehicle (UAV) platforms for operations. The applications of UAVs in fields such as military uses~\cite{konert2021military}, environmental monitoring~\cite{rohi2020autonomous}, and disaster relief~\cite{kucharczyk2021remote} are becoming increasingly widespread. The task of object detection, as an important area within UAV image processing, has attracted increasing attention from researchers~\cite{cao2021visdrone,wang2023highly,wang2021scaf,zhu2021tph,deng2020global,huang2022ufpmp}. High-mobility UAVs, known for their excellent maneuverability and flexibility, are capable of performing various tasks in complex environments~\cite{mohsan2022towards,wang2022high,agrawal2022novel}, especially in reconnaissance, search, and rescue operations. To ensure success in these operations, UAVs must maintain real-time and accurate perception of the environment.

However, compared to fixed or mobile observation platforms on the ground, UAVs in flight are more susceptible to disturbances from special environmental factors such as sudden changes in direction, airflow disturbances, and high-speed motion during observation and imaging. These factors often result in motion blur in the acquired images~\cite{huang2023uav,truong2020slimdeblurgan,zhu2023adaptive}. Particularly during highly maneuverable flight missions, these issues are more pronounced, where transient deviations in camera view can introduce global noise in images, adversely affecting UAV imaging quality and subsequent semantic interpretation tasks, potentially leading to mis-detections and omissions.

\begin{figure*}[h!]
	\centering
	\includegraphics[width=180mm]{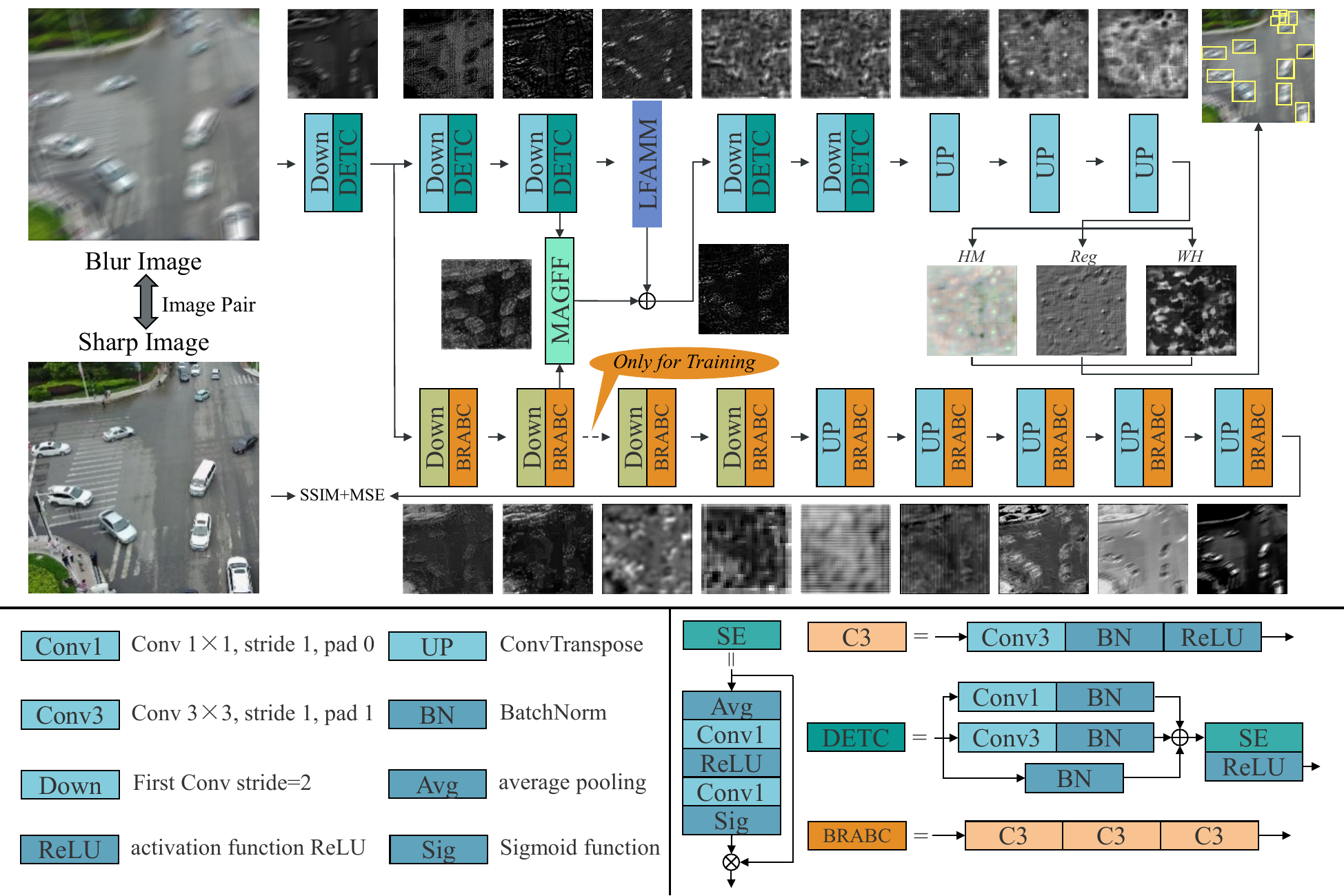}
	\caption{Overall architecture of the proposed DREB-Net. There are two main branches. One is the object detection main branch, and the other is the Blurry image Restoration Auxiliary Branch(BRAB). The BRAB consists of an encoder-decoder similar to U-Net. The object detection main branch obtains the feature maps of HM (HeatMap), WH (Width and Height) and Reg (Regression) after multiple convolutions and downsampling. In DREB-NET, we have added our innovative modules, MAGFF and LFAMM, which will be introduced in detail later.}
	\label{fig2}
\end{figure*}

Although advanced image restoration algorithms~\cite{qi2023e2nerf,shang2023joint,zhong2023blur} can improve the quality of blurry images to some extent and can be used as a pre-processing step for object detection images, these methods are usually based on complex deep learning models with large numbers of parameters and high computational complexity. Especially when processing large-scale or high-resolution images, these methods consume substantial computational resources and require lengthy processing times, severely limiting their use in real-time or near-real-time applications, particularly when deployed on UAV platforms.

To address the challenges of object detection in blurry images captured by high-mobility UAVs, we propose a novel object detection model named \textbf{DREB-Net} (Dual-stream Restoration Embedding Blur-feature Fusion Network). This model is specifically designed to solve the problem of object detection in blurry images. Specifically, DREB-Net enhances traditional detection architectures by incorporating a Blurry image Restoration Auxiliary Branch (\textbf{BRAB}). This model uses the Multi-level Attention-Guided Feature Fusion (MAGFF) module to fuse the shallow features from the BRAB with the features from the object detection branch. This design significantly boosts the robustness of DREB-Net when processing blurry images. The MAGFF module, which combines local and global attention mechanisms, assigns varying weights during the feature fusion process to optimize feature response and improve model training effectiveness. During the inference phase, to reduce computational complexity and increase processing speed, the deep feature extraction part of the BRAB will be removed. Additionally, DREB-Net introduces a Learnable Frequency domain Amplitude Modulation Module (\textbf{LFAMM}) in the early stage of feature extraction, which enhances the model's ability to handle motion blur by adjusting the frequency domain feature amplitudes. In the loss function, a combined loss of Mean Squared Error (MSE) and Structural Similarity Index Measure (SSIM) loss is specifically introduced for the BRAB to strengthen its capabilities of shallow feature extraction and provide higher quality features for subsequent object detection tasks. Extensive experiments conducted on UAV motion blur datasets demonstrate that DREB-Net has significant performance advantages in blurry image object detection. In general, The innovations of this paper can be summarized in the following main aspects:

\begin{itemize}
	\item Addressing the issue of detection failures caused by blurry images captured by high-mobility UAVs, we propose a novel object detection model that integrates a Blurry image Restoration Auxiliary Branch (\textbf{BRAB}), enhancing the processing ability of object detection for blurry images.
	\item A dual-stream feature fusion module named \textbf{MAGFF}(Multi-level Attention-Guided Feature Fusion) is designed, which dynamically adjusts feature fusion weights through local and global attention mechanisms, thereby adaptively optimizing feature response and effectively improving the model's ability to detect objects in blurry images.
	\item A Learnable Frequency domain Amplitude Modulation Module (\textbf{LFAMM}) is designed, utilizing frequency analysis technology to improve image quality, which is especially effective in processing motion blur images.
	\item A loss term for blurry image restoration is added to the loss function in the training stage to enhance the image restoration ability during the shallow feature extraction of the model, providing clearer feature information for object detection.
		
\end{itemize}
The remainder of this paper is organized as follows. Section~\ref{sec:introduction} provides a detailed introduction to the technical background and methods for object detection in blurry images of UAVs. Section~\ref{sec:relatedworks} provides an in-depth overview of the current research on object detection, blurry image restoration, and multitask learning. Section~\ref{sec:method} specifically details the method we propose for object detection in blurry images from high-mobility UAVs. Section~\ref{sec:experiments} provides information on the datasets, implementation settings, and experimental results. Finally, Section~\ref{sec:conclusion} concludes the paper.

\section{Related Works}
\label{sec:relatedworks}

\subsection{Object Detection}
Object detection technology is a fundamental area of research in unmanned aerial vehicle image processing tasks, with its development can be traced back to early image processing techniques, which evolved from traditional methods to advanced deep learning algorithms~\cite{zou2023object}. Prior to the rise of deep learning, object detection mainly relied on manual feature extraction and simple classifiers. For instance, the Viola-Jones face detection framework~\cite{viola2001rapid}, which utilized Haar features and Adaboost classifiers, achieved excellent detection results at the time. Subsequently, methods based on Support Vector Machines(SVM) and Histogram of Oriented Gradients(HOG) features were also widely used for pedestrian detection~\cite{dalal2005histograms}. However, these methods often underperformed in dealing with complex backgrounds or diverse environments typical in drone images. With the advent of deep learning technologies, convolutional neural network-based object detection models have become the mainstream, fundamentally transforming the application of object detection techniques in UAV image processing.

The object detection methods based on convolutional neural networks can be categorized into two-stage detection strategy and single-stage detection strategy~\cite{zhao2019object}. Two-stage object detection algorithms such as R-CNN~\cite{girshick2014rich}, and its variants Fast R-CNN~\cite{girshick2015fast} and Faster R-CNN~\cite{ren2015faster} have significantly improved detection accuracy and robustness by learning features from a large number of datasets. Faster R-CNN, by introducing the Region Proposal Network (RPN), has made the extraction of candidate regions more efficient. In the object detection task of UAV images or remote sensing images, methods such as ~\cite{li2018r} can effectively recognize and locate vehicles in different directions by integrating rotation-invariant features and region proposal network, and ~\cite{wang2021normalized} used the Normalized Wasserstein Distance (NWD) for detecting small objects, both of which have achieved good results.

With the increasing demand for real-time object detection, single-stage detection algorithms, such as the YOLO series~\cite{redmon2016you, redmon2018yolov3, chen2021you, ge2021yolox,  wang2024yolov9}, the SSD series~\cite{liu2016ssd} and FSAF~\cite{zhu2019feature}, are receiving more and more attention. These algorithms abandon the traditional two-step proposal and classification process, predicting bounding boxes and categories in a single forward propagation, significantly accelerating inference speed. In particular, the YOLO series object detection algorithms have continuously optimized across multiple iterations, with significant improvements in detection speed and accuracy. In addition, improvements in loss functions such as Focal Loss~\cite{lin2017focal} solve the problem of imbalance between foreground and background samples and have improved the performance of the one-stage object detection algorithm. In the task of object detection in the UAV scene, methods such as~\cite{tijtgat2017embedded, yu2021towards, mandal2019avdnet, qin2020specially, lu2019gated} have achieved excellent results. In recent years, with the development of Vision Transformer, many methods based on Vision Transformer have also been widely applied in object detection tasks, such as DETR~\cite{carion2020end}, DINO~\cite{zhang2022dino}, etc.

Furthermore, object detection algorithms can also be divided into anchor-based methods and anchor-free methods. The anchor-based methods predict the position and size of objects through predefined anchors, which are a series of preset rectangular boxes at different scales and aspect ratios. The model needs to learn to adjust these rectangular boxes to match the actual object best. Classic anchor-based methods include Faster R-CNN~\cite{ren2015faster}, SSD~\cite{liu2016ssd}, and YOLOv3~\cite{redmon2018yolov3}. The anchor-free methods eliminate traditional anchor boxes, making the object detection process more concise and intuitive, by directly predicting the key points or center points of objects and determining their boundaries from these points. Popular anchor-free detection algorithms include CornerNet~\cite{law2018cornernet}, CenterNet~\cite{zhou2019objects}, FCOS~\cite{tian2022fully}, VFNet~\cite{zhang2021varifocalnet}, CentripetalNet~\cite{dong2020centripetalnet}, CentripetalNet~\cite{zhou2019objects}, and recent iterations of YOLO such as YOLOv9~\cite{wang2024yolov9}.

\subsection{Blurry Image Restoration}
Cameras capture and record light information through a series of optical and electronic components. The working principle of cameras is to record scene information through a photosensitive surface during the exposure time. A typical camera consists of a lens, an aperture, a shutter, and photosensitive components. The function of the lens is to focus, the aperture is used to control the amount of light entering the camera, and the shutter determines the duration of light impacts the photosensitive element. Once the light is focused through the lens, it forms a real image on the image sensor, which converts the received light signal into an electrical signal, and then converts it into a digital image through A/D conversion. During this process, image quality is influenced by many factors, such as lighting conditions, camera settings, and external environments~\cite{le2017survey}.

During the camera imaging process, light is focused on the photosensitive element through the lens. Ideally, if both the camera and the subject remain stationary, the light is directly mapped to the corresponding position on the photosensitive element, forming a clear image. However, suppose the camera or the object moves during the exposure time. In that case, the projection path of the light will change, extending from one pixel to multiple pixels, forming a trajectory and causing blurry stripes on the image. The motion patterns of drones are more complex than those of general photographic equipment, including translations, rotations, and vibrations. These factors collectively complicate the imaging process, thereby increasing the complexity of blur.

Blur image restoration technology is an important research direction that is continuously evolving in the field of image processing, which is commonly referred to as deblur. Deblur techniques can be categorized into blind deblur methods and non-blind deblur methods~\cite{zhang2022deep}. Blind deblur methods do not require prior knowledge about the blur kernel, while non-blind deblur methods require information about the blur kernel. With advancements from traditional to deep learning methods, deblur technology has made significant progress~\cite{li2022survey}. Traditional deblur methods are mainly based on statistical characteristics of images, such as Wiener filtering~\cite{robinson1967principles} and the Lucy-Richardson algorithm~\cite{singh2007adaptively}. However, these methods rely on complex optimization processes, have high computational costs, and are often limited in effectiveness in dealing with practical complex blurs. With the rise of deep learning, deblur methods based on convolutional neural networks have shown excellent performance. Deep learning methods can automatically learn effective features for blur image restoration from a large amount of data. The pioneering work using deep learning methods for blur image restoration tasks can be traced back to ~\cite{sun2015learning} in 2014, which effectively restored images by directly learning the mapping relationship from blurred to clear images. Subsequently, methods such as~\cite{nah2017deep, tao2018scale} used deep convolutional neural networks to restore blurred images by directly learning the mapping from blur to clear images. These methods typically involve multi-layer networks, and each layer is designed to enhance local details of the image incrementally. Later, DeblurGAN ~\cite{kupyn2018deblurgan} and its improved version DeblurGANv2 \cite{kupyn2019deblurgan} utilized Generative Adversarial Networks(GAN), employing adversarial training to achieve the restoration of motion blur images, which involves a generator creating deblurred images and a discriminator distinguishing between generated and real clear images, significantly improving the quality and naturalness of the restored images. Recently, Transformer-based methods have also been applied to blur image restoration tasks, such as\cite{wang2022uformer, tsai2022stripformer, zamir2022restormer}, which can effectively deal with complex blur in images by using the transformer's powerful global dependency capture ability, especially on large-size images. However, these methods generally are complex in model architecture, with a large number of parameters and computational delays, and often are unsuitable for real-time image processing in UAVs.

\subsection{Multi-task Learning}
Multi-task Learning is a significant research field in machine learning, whose core idea is to enhance model generalization, improve data efficiency, and boost learning efficiency by simultaneously training multiple related tasks, as well as facilitating knowledge transfer between tasks~\cite{zhang2021survey}. The theoretical foundation of multi-task learning can be traced back to the 1990s when researchers began to explore whether joint training could improve the learning efficiency and performance of several related tasks. With the development and popularization of deep learning technologies, the methods and applications of multi-task learning have been significantly expanded and deepened, especially when dealing with high-dimensional data and complex model structures, where multi-task learning has shown unique advantages.

For example, MultiNet~\cite{teichmann2018multinet} uses a shared convolutional architecture to handle three key tasks: semantic segmentation, object detection, and lane line detection in traffic monitoring systems. This model optimizes the performance of each task by using task-specific branches while also improving data processing efficiency and model generalization ability by sharing the underlying layers. DeepFace~\cite{wang2017multi} employs a deep convolutional network to achieve multi-task learning for facial recognition, gender classification, and emotion recognition. Through end-to-end training, the model can effectively share features among tasks, significantly improving processing speed and accuracy. HyperFace~\cite{ranjan2017hyperface} utilizes a single CNN model with a branch structure to simultaneously perform tasks of face detection, facial landmark localization, gender recognition, and age estimation. This integrated multitasking approach not only optimizes the use of computational resources but also improves the execution efficiency and accuracy of various tasks.

However, multi-task learning also faces several challenges, such as task conflict, where the optimal solutions of different tasks may contradict each other, potentially leading to a decline in model performance. Moreover, designing effective task weights and loss functions, as well as balancing the learning speed and priority among tasks, are crucial for implementing successful multi-task learning.

\section{Methodology}
\label{sec:method}
In this section, we provide a detailed introduction to the method we propose. First, we introduce the overall architecture of the DREB-Net, designed specifically for object detection in blurry images of UAVs. This architecture comprises an object detection network branch, a blurry image restoration auxiliary branch, the MAGFF module, and the LFAMM module. Next, we will elaborate on the functions and roles of these components. In addition, in order to optimize the feature extraction ability of the model during the downsampling phase, we designed two task-driven loss functions. In order to illustrate the workflow of our method more intuitively, we provide detailed steps of the training process in Algorithm 1.

\subsection{Overall Architecture}
The overall architecture of DREB-Net is shown in Figure~\ref{fig2}. This network takes a single frame of image data as input and is specially designed to efficiently handle object detection tasks in blurry images captured by UAVs. The network consists of an object detection branch and a blurry image restoration auxiliary branch. The object detection branch captures image details and deep semantic information by progressively extracting image features. The input image first undergoes multiple DET-Conv operations and downsampling steps, reducing the feature map size to 1/32 of the original size. Then, it is followed by three deconvolution operations, which enlarge the feature map size is enlarged to 1/4 of the original size. Subsequently, the enlarged feature maps are fed into the HM (HeatMap), WH (Width and Height), and Reg (Regression) branches for further processing. The HM branch generates heatmaps using a Gaussian kernel to predict the position of the center point of each object category in the image. Each channel of the heatmap corresponds to an object category, with values indicating the confidence that a position is the center of an object. The WH branch is used to predict the width and height of each object in the image, and these predicted values, combined with the center points determined by the heatmap, jointly define the bounding boxes of the objects. The Reg branch is used to refine and adjust the position of the center points to improve the accuracy of pixel-level prediction.

The blurry image restoration auxiliary branch adopts an architecture similar to U-Net, which is a widely used encoder-decoder network in image restoration tasks. In this branch, the image detail and size are progressively restored through BRAB-Conv, downsampling, and upsampling operations. The encoder captures deep features of images through continuous convolution and downsampling operations, while the decoder restores image size and details through upsampling and convolution operations, and fuses deep and shallow features from the encoder through a feature aggregation module. This auxiliary branch provides more accurate information to improve the accuracy of object detection.

During the training phase, the input image is first processed by both the object detection branch and the BRAB to extract shallow features respectively. Subsequently, the MAGFF module is used to fuse these shallow features from both branches adaptively, and the fused features are then fed into the object detection network for further computation, while the deep features of the BRAB continue to operate using their features only. Additionally, the LFAMM module is introduced to enhance the model’s ability to handle motion blur situations by adjusting the amplitude of frequency domain features. A detailed introduction of the MAGFF modules and LFAMM modules will be provided later.

\begin{algorithm}[t!]
	\caption{Training Process of DREB-Net}
	\begin{algorithmic}[1]
		\State \textbf{Input:} 
		\State \hspace{\algorithmicindent} Blur images $X_{blur}$, 
		\State \hspace{\algorithmicindent} Sharp images $X_{sharp}$, 
		\State \hspace{\algorithmicindent} Object detection annotations $Y$
		\State \textbf{Output:} 
		\State \hspace{\algorithmicindent} Object detection results:
		\State \hspace{\algorithmicindent} category $c$, center coordinates $(x_c, y_c)$, width $w$, height $h$, confidence $conf$
		\State \textbf{Initialize:} 
		\State \hspace{\algorithmicindent} object DETection Branch (DET branch), 
		\State \hspace{\algorithmicindent} Blurry image Restoration Auxiliary Branch (BRAB)
		\State \hspace{\algorithmicindent} Learnable Frequency domain Amplitude Modulation Module (LFAMM)
		\For{epoch = 1 to 200}
		\For{each batch $(x, y)$ in training dataset}
		\State $f_{\text{det\_shallow}} \gets \text{DET\_shallow}(x)$  
		\State $f_{\text{brab\_shallow}} \gets \text{BRAB\_shallow}(x)$  
		\State $f_{\text{magff}} \gets \text{MAGFF}(f_{\text{det\_shallow}}, f_{\text{brab\_shallow}})$ 
		\State $f_{\text{lfamm}} \gets \text{LFAMM}(f_{\text{det\_shallow}})$ 
		\State $f_{\text{fused}} \gets \text{Add}(f_{\text{lfamm}}, f_{\text{brab\_shallow}})$ 
		\If{epoch $\leq$ 100}
		\State $f_{\text{det\_deep}} \gets \text{DET\_deep}(f_{\text{fused}})$  
		\State $f_{\text{brab\_deep}} \gets \text{BRAB\_deep}(f_{\text{brab\_shallow}})$  
		\State $L_{\text{det}} \gets w_{hm} \times L_{\text{HM}}(f_{\text{det\_deep}}, y) + w_{wh} \times L_{\text{WH}}(f_{\text{det\_deep}}, y) + w_{reg} \times L_{\text{Reg}}(f_{\text{det\_deep}}, y)$
		\State $L_{\text{brab}} \gets w_{mse} \times  L_{\text{MSE}}(f_{\text{brab\_deep}}, y) + w_{ssim} \times  L_{\text{SSIM}}(f_{\text{brab\_deep}}, y)$
		\State $L_{\text{total}} \gets L_{\text{det}} + L_{\text{brab}}$
		\Else
		\State {// Deep computation of BRAB is removed.}
		\State $L_{\text{det}} \gets w_{hm} \times L_{\text{HM}}(f_{\text{det\_deep}}, y) + w_{wh} \times L_{\text{WH}}(f_{\text{det\_deep}}, y) + w_{reg} \times L_{\text{Reg}}(f_{\text{det\_deep}}, y)$
		\EndIf
		\State $\nabla \theta \gets \partial L_{\text{total}} / \partial \theta$  
		\State $\theta \gets \theta - \eta \times \nabla \theta$  
		\EndFor
		\EndFor
	\end{algorithmic}
\end{algorithm}

It is important to note that during the inference phase of the model, to reduce dependence on computational resources and improve inference speed, the parts of the BRAB that are not related to feature fusion will be removed. This design enables DREB-Net to not only effectively locate and classify objects but also to handle image blur caused by camera shaking or rapid movement, thereby demonstrating excellent performance for object detection in blurry images. The process of our method can be illustrated by Algorithm 1.

\subsection{MAGFF module}
\begin{figure}[tb!]
	\centering
	\includegraphics[width=81mm]{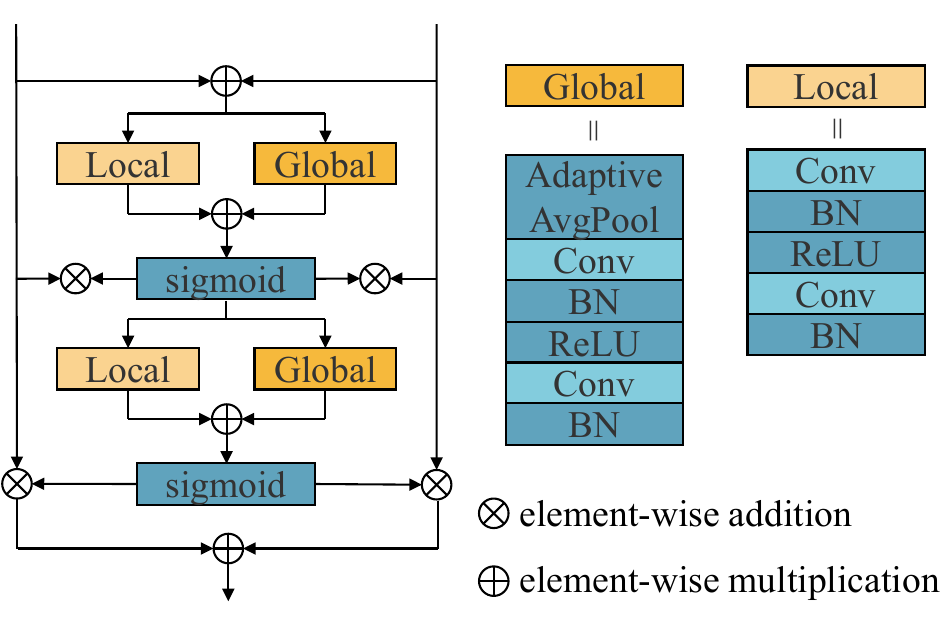}
	\caption{MGAFF module. It consists of local attention submodules and global attention submodules, which process input features through different paths and select and fuse features through the sigmoid function. It is used to optimize the integration of global information and improve the detail perception ability of features when the object detection main branch and the BRAB perform feature fusion.}
	\label{fig3}
\end{figure}

The MAGFF(Multi-level Attention-Guided Feature Fusion) module is an adaptive feature fusion module designed for fusing shallow features. Its main purpose is to effectively fuse shallow features derived from the BRAB with those extracted by the object detection network, thereby improving the performance of the detection results. The MAGFF module utilizes a combination of local and global attention mechanisms to improve the effect of feature fusion and the selection process. The module is illustrated in Figure~\ref{fig3}.

Specifically, the MAGFF module consists of two layers of attention mechanisms, each of which contains a local attention submodule and a global attention submodule. These two attention mechanisms process input features through different paths and complete the feature selection and fusion using the sigmoid function.

The local attention mechanism aims to capture detailed features in the image, which is achieved through a series of convolutional layers and batch normalization layers. This process can be expressed as:
\begin{equation}
	\begin{split}
		L(x)=f_{BN2}\circ f_{C2}\circ f_{ReLU}\circ f_{BN1}\circ f_{C1}(x),   \label{Eq:1}
	\end{split}
\end{equation}
where $f_{C1}$ and $f_{C2}$ denote two 1×1 convolution operations applied to the input $x$, $f_{BN1}$ and $f_{BN2}$ represent the corresponding batch normalization operations, $f_{ReLU}$ represents the ReLU activation function, $\circ$ represents function composition, where the output of one function is used as the input of another function.

The global attention mechanism captures the global contextual information of the input features through an adaptive average pooling layer, and then processed through convolution and batch normalization layers to achieve global feature adjustment:
\begin{equation}
	\begin{split}
		G(x)=f_{BN2}\circ f_{C2}\circ f_{ReLU}\circ f_{BN1}\circ f_{C1}\circ f_{AAP}(x),   \label{Eq:2}
	\end{split}
\end{equation}
where $f_{AAP}$ represents the adaptive average pooling operation applied to the input $x$.

The outputs of local and global attention will be added together, and the sigmoid function will be used to generate fusion weights. These weights are then applied to fuse and adjust the input features:
\begin{equation}
	\begin{split}
		x_{\text{out}} = &\sigma(Att_L(x_1) + Att_G(x_1)) \cdot x_1 \\
		& + (1 - \sigma(Att_L(x_1) + Att_G(x_1))) \cdot x_2,   \label{Eq:3}
	\end{split}
\end{equation}
where $\sigma$ represents the Sigmoid function, and $x_1$ and $x_2$ represent inputs from two network branches, respectively.

To further enhance feature expression, we reapply local and global attention mechanisms to the fused features again, and adjust the weights to get the final output:
\begin{equation}
	\begin{split}
		x_{\text{final}} = & \sigma(Att_L(x_{\text{out}}) + Att_G(x_{\text{out}})) \cdot x_{\text{out}} \\
		& + (1 - \sigma(Att_L(x_{\text{out}}) + Att_G(x_{\text{out}}))) \cdot x_2,   \label{Eq:4}
	\end{split}
\end{equation}
where $x_{final}$ represents the final output features.

The MAGFF module effectively enhances the expressiveness and relevance of the input features by combining local and global attention mechanisms, which not only improves the ability of detail perception but also optimizes the integration of global information, thereby significantly improving the performance of object detection of blurry images. Through the iterative feature fusion process, the module is able to adapt to different image contents, providing more accurate and robust feature expressions, which is crucial for improving the accuracy of object detection in complex environments.

\subsection{LFAMM module}
\begin{figure}[tb!]
	\centering
	\includegraphics[width=85mm]{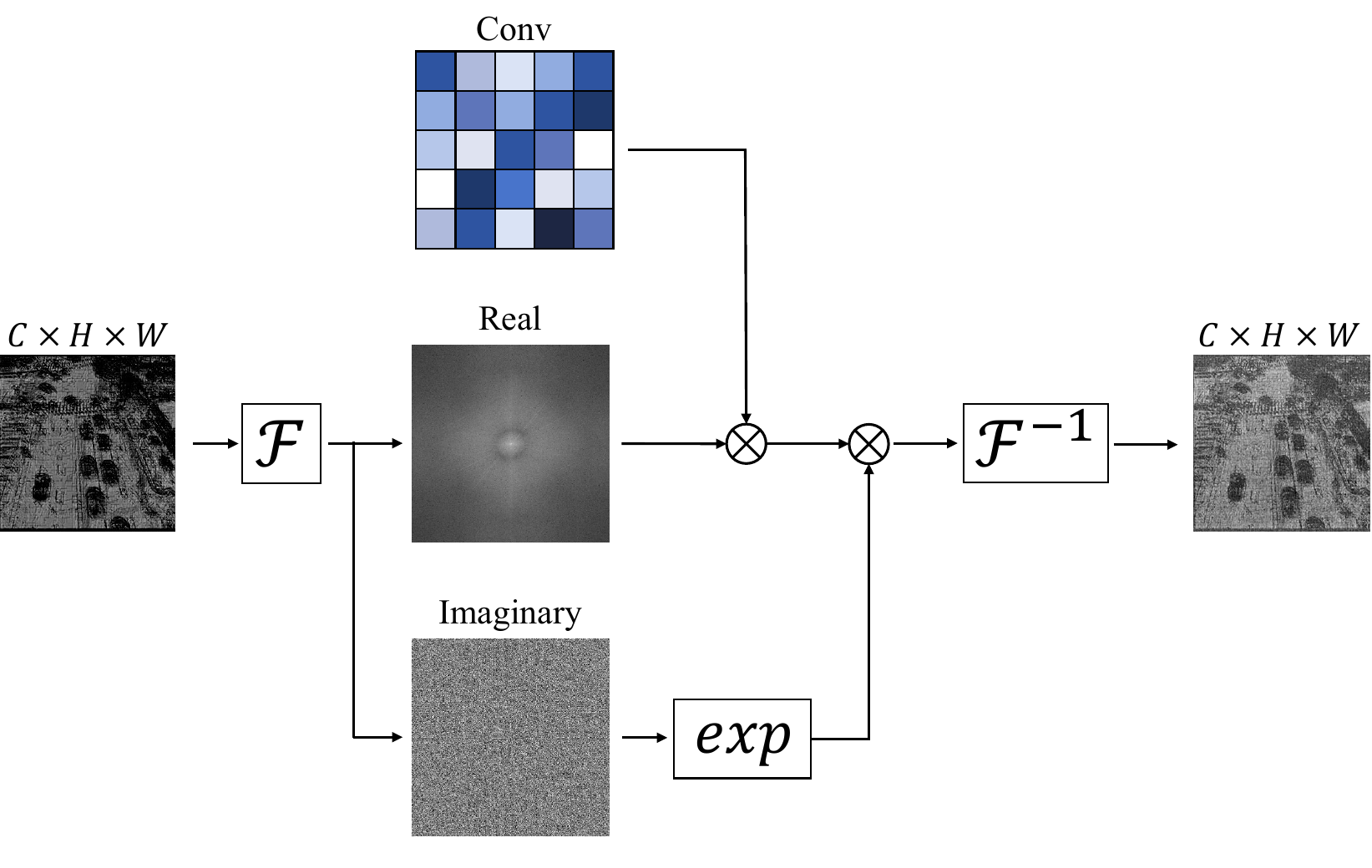}
	\caption{LFAMM module. It relies on frequency domain processing to enhance the feature expression ability of input images. It uses the Real Fast Fourier Transform (RFFT) and its inverse (IRFFT) and convolution operation to preserve the spatial information of the image while adjusting its frequency components to enhance the details and quality of the image.}
	\label{fig4}
\end{figure}

In order to improve the accuracy and robustness of object detection in motion blur images, we designed a Learnable Frequency domain Amplitude Modulation module, named LFAMM in DREB-Net. This module adjusts the frequency components of the image while retaining its spatial information to enhance the details and quality of the image. It relies on frequency domain processing to enhance the feature expression ability of input images, as shown in Figure~\ref{fig4}.

The module receives images of specific width and height as input. It uses the Real Fast Fourier Transform (RFFT) and its inverse (IRFFT), combined with a learnable frequency domain filter, to preprocess the input image. The specific implementation process is as follows.

First, assume that the input image is represented in the spatial domain as $X \in \mathbb{R}^{C \times H \times W}$, where $C$, $H$, and $W$ denote the number of channels, height, and width of the image, respectively. Through the Real Fast Fourier Transform (RFFT), the image is transformed from the spatial domain to the frequency domain:
\begin{equation}
	\begin{split}
		F(X) = \text{RFFT}(X) \in \mathbb{R}^{C \times H \times \left(\lfloor \frac{W}{2} \rfloor + 1\right)},   \label{Eq:5}
	\end{split}
\end{equation}
where $F(X)$ denotes the complex representation of the image in the frequency domain, with each element containing both amplitude and phase components.

In order to adjust the frequency components of the image, a learnable frequency domain filter $W \in \mathbb{R}^{C \times H \times \left(\lfloor \frac{W}{2} \rfloor + 1\right)}$ is designed. This filter performs a multiplication operation with the amplitude component of the image frequency domain to enhance or suppress specific frequency components. The filtering process can be represented as:
\begin{equation}
	\begin{split}
		A = |F(X)| \cdot W,   \label{Eq:6}
	\end{split}
\end{equation}
where $|F(X)|$ denotes the amplitude part of $F(X)$, and $A$ denotes the adjusted amplitude.

After adjusting the amplitude, the module maintains its phase unchanged and needs to reconstruct the representation in the frequency domain in order to convert it back to the spatial domain. The reconstructed frequency domain image $\tilde{F}(X)$ is defined as:
\begin{equation}
	\begin{split}
		\tilde{F}(X) = A exp(i\cdot arg(F(X))),   \label{Eq:7}
	\end{split}
\end{equation}
where $arg(F(X))$ denotes the phase component of $F(X)$, and $i$ is the imaginary unit.

Finally, through the Real Inverse Fast Fourier Transform (IRFFT), the adjusted frequency domain image $\tilde{F}(X)$ is converted back to the spatial domain:
\begin{equation}
	\begin{split}
		\tilde{X} = \text{IRFFT}(\tilde{F}(X)) \in \mathbb{R}^{C \times H \times W},   \label{Eq:8}
	\end{split}
\end{equation}

The LFAMM module completes the entire preprocessing process of the image from input to output, significantly improving the flexibility of frequency domain adjustments and sensitivity to image details through the learnable frequency processing module, thereby enhancing the detection capabilities for objects within the image. The LFAMM module is not only suitable for object detection in blurry images but also opens new technical paths for other image processing tasks such as noise suppression, detail enhancement, etc.

\subsection{Loss function}
Next, we provide a detailed description of the loss function designed for DREB-Net, which combines several key aspects of the accuracy and robustness of object detection in blurry images. The loss function consists of the following parts: losses for heatmaps, width and height of bounding boxes, offsets for object detection branch, and MSE, SSIM losses for BRAB. The specific implementation is as follows.

1) Heatmap Loss: The heatmap loss uses Focal Loss, primarily to solve the problem of imbalance between background and foreground categories in object detection. Focal Loss effectively handles category imbalance by adjusting the weights of easily classified samples. The formulation is as follows:
\begin{equation}
	\begin{split}
		L_{hm} = \frac{1}{N} \sum_{i=1}^{N} (1 - p_i)^{\gamma} \log(p_i),   \label{Eq:9}
	\end{split}
\end{equation}
where $p_i$ is the predicted probability of the heatmap, $\gamma$ is the focusing parameter that adjusts the weight of easily classified samples, and $N$ is the batch size.

2) Width and Height Loss: The width and height loss employs L1 Loss to accurately predict the bounding box size of the object, ensuring that the predicted width and height minimize the difference between the true values:
\begin{equation}
	\begin{split}
		L_{wh} = \frac{1}{M} \sum_{j=1}^{M} |wh_j - \hat{wh}_j|,   \label{Eq:10}
	\end{split}
\end{equation}
where $wh_j$ represents the actual width and height, $\hat{wh}_j$ represents the predicted width and height, and $M$ is the number of positive samples.

3) Offset Loss: The offset loss also uses L1 Loss to correct the predicted position of the object center point and reduce the positioning error of the center point:
\begin{equation}
	\begin{split}
		L_{off} = \frac{1}{M} \sum_{k=1}^{M} |off_k - \hat{off}_k|,   \label{Eq:11}
	\end{split}
\end{equation}
where $off_k$ and $\hat{off}_k$ represent the actual and predicted center point offsets, respectively.

4) MSE Loss: Mean Squared Error is a loss used for blurry image restoration. In DREB-Net, it is used together with the structural similarity index (SSIM Loss) to evaluate the effectiveness of blurry image restoration comprehensively. The combination of these two losses helps the model optimize the overall visual quality while preserving image details. MSE loss measures the pixel-level difference between the restored image and the clear real image. It is defined by calculating the mean of the squared differences between the pixel values of the predicted image and the real image, ensuring the accurate restoration of the image content. The formula is as follows:
\begin{equation}
	\begin{split}
		L_{MSE} = \frac1N\sum_{i=1}^N(I_{sharp,i}-I_{res,i})^2,   \label{Eq:12}
	\end{split}
\end{equation}
where $I_{sharp}$ is the clear, real image, $I_{res}$ is the restored image, and $N$ is the total number of pixels in the image.

\begin{figure*}[htbp]
	\centering
	\begin{subfigure}{0.48\textwidth}
		\includegraphics[width=\linewidth]{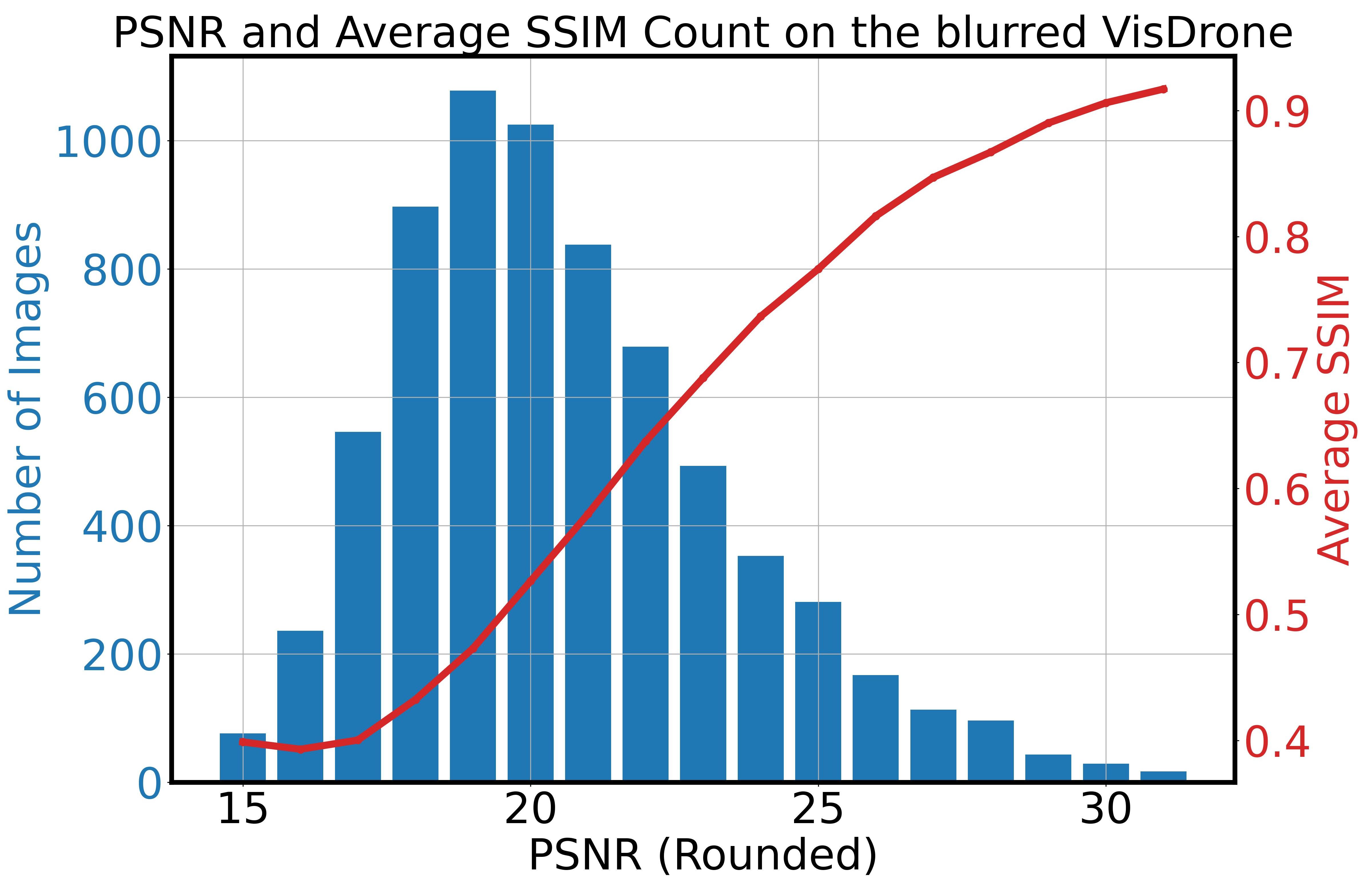}
		\caption*{(a)}
	\end{subfigure}
	\hfill
	\begin{subfigure}{0.48\textwidth}
		\includegraphics[width=\linewidth]{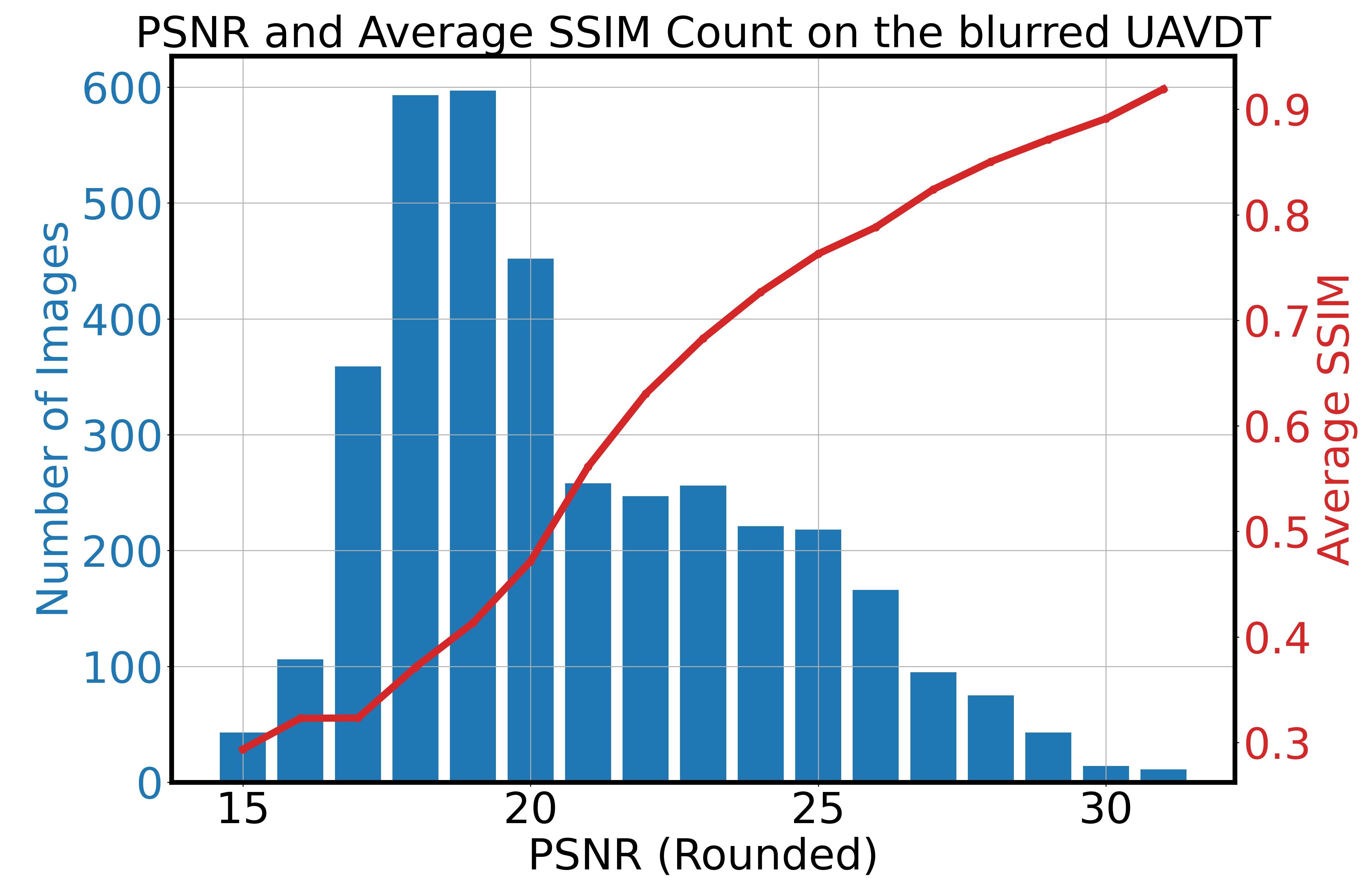}
		\caption*{(b)}
	\end{subfigure}
	\caption{Statistics histogram of image count by PSNR and average SSIM per PSNR for the blurred image datasets. (a) on the blurred VisDrone-2019-DET dataset. (b) on the blurred UAVDT dataset.}
	\label{fig:PSNR}
\end{figure*}

5) SSIM(Structural Similarity) Loss: The SSIM loss measures the similarity in visual structure, brightness, and contrast between the restored image and the clear image. SSIM loss is a more comprehensive loss metric that considers not only pixel differences but also the structural information and perceptual quality of the image, helping to maintain the structural integrity and visual effects of the image:
\begin{equation}
	\begin{split}
		L_{SSIM} = 1- \frac{(2\mu_x\mu_y+c_1)(2\sigma_{xy}+c_2)}{(\mu_x^2+\mu_y^2+c_1)(\sigma_x^2+\sigma_y^2+c_2)},   \label{Eq:13}
	\end{split}
\end{equation}
where $x$ and $y$ are the window areas of the two images to be compared. $\mu_x$ and $\mu_y$ are the means of $x$ and $y$ respectively. $\sigma_x^2$ and $\sigma_y^2$ are the variances of $x$ and $y$ respectively. $\sigma_{xy}$ is the covariance of $x$ and $y$. $c_1=(k_1L)^2$, $c_2=(k_2L)^2$ are small constants for stability, and $L$ is the dynamic range of the data, $k_1=0.01$ and $k_2=0.03$ are the default values.

\textbf{Overall Loss}: The overall loss is a weighted sum of the above losses, including heatmap loss, width and height loss, offset loss, MSE Loss and SSIM loss. The weights of each loss component can be adjusted according to specific task requirements to optimize the performance of the model in specific applications:

\begin{equation}
	\begin{split}
		L = & w_{hm} \cdot L_{hm} \\
		& + w_{wh} \cdot L_{wh} \\
		& + w_{off} \cdot L_{off} \\
		& + w_{MSE} \cdot L_{MSE} \\
		& + w_{SSIM} \cdot L_{SSIM},   \label{Eq:14}
	\end{split}
\end{equation}
where $w_{hm}$,$w_{wh}$,$w_{off}$,$w_{deblur}$,$w_{SSIM}$ are the weights corresponding to each loss component.

By meticulously designing the loss function and adjusting the weights of different loss terms during the training phase, the model is able to handle object detection tasks in blurry images, especially in terms of object localization and recognition accuracy. The design of those loss functions not only considers the complexity of the object detection task but also the importance of image quality, thereby optimizing model training and maximizing effectiveness.

\section{Experiments}
\label{sec:experiments}
In this section, we first introduce the datasets and experimental details used for the experiments. Then, we compare the object detection performance of the method we proposed in this paper with other excellent methods on blurred drone images. In addition, extensive ablation experiments are conducted to evaluate the effectiveness of our method. Finally, the feasibility of DREB-Net in actual blurry image object detection tasks is discussed.

\subsection{Dataset Description}
Firstly, we provide a detailed introduction to the datasets used in the experiments, as well as the method we used to generate blurred images. We chose the VisDrone-2019-DET dataset and UAVDT dataset as the basis for our experiment, which are widely used in UAV image and large-scale ground object detection research.

VisDrone-2019-DET dataset: The VisDrone-2019-DET dataset consists of images captured by drones from different cities and environments. The official data includes 6471 training images and 548 validation images, with precise object locations and category annotations. These images are highly diverse and complex, making them an ideal choice for evaluating object detection algorithms in drone images. We used the officially designated training and validation sets.

UAVDT dataset: The UAVDT dataset is a large-scale vehicle detection and tracking dataset designated by the University of the Chinese Academy of Sciences in 2018. It contains 100 video sequences, covering various weather conditions, occlusions, and flight altitudes. Among them, the 50 video sequences for multi-object tracking have detailed annotations of the location and category of interest objects in each frame. The objects in the dataset are mainly moving vehicles, including cars, trucks and buses, which vary greatly in size, speed, and direction. The dataset shows a variety of common scenes, including squares, main roads, toll booths, highways, and intersections. It can be used for various tasks such as vehicle detection, single-vehicle tracking, and multi-vehicle tracking.

\begin{table*}[htbp!]
	\centering
	\caption{Test Results on The Blurred VisDrone-2019-DET Dataset. DREB-Net\_tiny Model Means That SE Blocks Have Been Removed from DETC in DREB-Net.}
	\label{table1}
	\setlength{\tabcolsep}{5pt}  
	\begin{tabular}{ccccccccccccccc}
		\hline
		\multirow{2}{*}{{Models}} & \multicolumn{4}{c}{$AP_{50}$} & \multicolumn{4}{c}{$AR_{50}$} & \multirow{2}{*}{$mAP_{s}$} & \multirow{2}{*}{$mAP_{m}$} & \multirow{2}{*}{$mAP_{l}$} & \multirow{2}{*}{$mAP_{50}$} & \multirow{2}{*}{$mAR_{50}$}  & \multirow{2}{*}{FPS}\\ \cline{2-9}
		& {people} & {car} & {truck} & {bus} & {people} & {car} & {truck} & {bus} & & & & & & \\ \hline
		{Faster R-CNN~\cite{ren2015faster}} & 0.083 & 0.487 & 0.159 & 0.217 & 0.331 & 0.414 & 0.307 & 0.392 & 0.063 & 0.388 & 0.615 & 0.237 & 0.347 &  16.35 \\ \hline
		{RetinaNet~\cite{lin2017focal}} & 0.053 & 0.412 & 0.062 & 0.067 & 0.253 & 0.380 & 0.176 & 0.197 & 0.029 & 0.272 & 0.535 & 0.148 & 0.257  & 17.08 \\ \hline
		{CornerNet~\cite{law2018cornernet}} & 0.087 & 0.326 & 0.094 & 0.214 & 0.259 & 0.326 & 0.189 & 0.278 & 0.093 & 0.294 & 0.308 & 0.180 & 0.292 & 2.91  \\ \hline
		{CenterNet~\cite{zhou2019objects}} & 0.113 & 0.502 & 0.176 & 0.304 & 0.321 & 0.404 & 0.300 & 0.421 & 0.088 & 0.440 & 0.779 & 0.274 & 0.378  & 26.85 \\ \hline
		{FCOS~\cite{tian2022fully}} & 0.074 & 0.433 & 0.102 & 0.138 & 0.321 & 0.400 & 0.231 & 0.214 & 0.048 & 0.303 & 0.632 & 0.187 & 0.284  & 17.40  \\ \hline
		{FSAF~\cite{zhu2019feature}} & 0.097 & 0.455 & 0.112 & 0.108 & 0.303 & 0.385 & 0.215 & 0.165 & 0.069 & 0.302 & 0.612 & 0.193 & 0.294  & 17.55  \\ \hline
		{YOLOX~\cite{ge2021yolox}} & 0.115 & 0.523 & 0.122 & 0.251 & 0.374 & 0.427 & 0.272 & 0.385 & 0.093 & 0.400 & 0.696 & 0.253 & 0.386  & 37.69  \\ \hline
		{YOLOF~\cite{chen2021you}} & 0.071 & 0.409 & 0.115 & 0.238 & 0.336 & 0.403 & 0.256 & 0.382 & 0.052 & 0.349 & 0.681 & 0.208 & 0.320  &  29.48 \\ \hline
		{VFNet~\cite{zhang2021varifocalnet}} & 0.157 & 0.551 & 0.119 & 0.146 & 0.369 & 0.420 & 0.303 & 0.282 & 0.118 & 0.385 & 0.605 & 0.243 & 0.363  &  14.65 \\ \hline
		{CentripetalNet~\cite{dong2020centripetalnet}} & 0.170 & 0.568 & 0.145 & 0.281 & 0.359 & 0.418 & 0.292 & 0.401 & 0.139 & 0.461 & 0.695 & 0.291 & 0.387  & 2.37  \\ \hline
		{DINO~\cite{zhang2022dino}} & 0.174 & 0.563 & 0.116 & 0.270 & 0.362 & 0.416 & 0.231 & 0.443 & 0.144 & 0.430 & 0.614 & 0.281 & 0.423 & 7.92  \\ \hline
		{DREB-Net\_tiny} & 0.193 & 0.589 & 0.198 & 0.317 & 0.381 & 0.429 & 0.296 & 0.434 & 0.142 & 0.496 & 0.800 & 0.320 & 0.427 & 31.85  \\ \hline
		\textbf{DREB-Net} & \textbf{0.213} & \textbf{0.599} & \textbf{0.226} & \textbf{0.337} & \textbf{0.388} & \textbf{0.430} & \textbf{0.342} & \textbf{0.469} & \textbf{0.161} & \textbf{0.523} & \textbf{0.792} & \textbf{0.344} & \textbf{0.448}  & 11.57  \\ \hline		
	\end{tabular}
\end{table*}

In order to examine the ability to process blurry images of DREB-Net in experiments, we generate blurred images according to the method in ~\cite{truong2020slimdeblurgan}. Specifically, we use the original clear images as input and generate the corresponding blurred images through the above method. This method can simulate dynamic blur caused by rapid movement, making the generated image closer to those captured by drones or other fast-moving camera devices in natural environments. However, despite the fact that the image has been blurred, we still use the object detection labels of the original clear image as annotations during training and testing. The purpose of this is to ensure the consistency and accuracy of annotations while evaluating the performance and robustness of the model when processing blurry images. To quantify the quality of generated images, we calculated the Structural Similarity Index (SSIM) and Peak Signal to Noise Ratio (PSNR). As shown in Figure~\ref{fig:PSNR}, we provide statistical graphs of SSIM and PSNR values for all images in the two datasets. These indicators reflect the similarity between the generated images and the original images in terms of visual quality and information retention, providing a validation basis for subsequent object detection algorithms.

Through these experimental settings, we are able to comprehensively evaluate the effectiveness and accuracy of the proposed object detection method in processing images of different types and degrees of blur. It not only helps to validate the practicality of the model, but also explores possible paths for further improving the model to adapt to various complex application scenarios.

\subsection{Implementation Details}
In this paper, all experiments are conducted using the PyTorch framework. The operating environment used in the experiment is equipped with four NVIDIA Quadro RTX 6000 GPUs, for Intel Xeon Silver 4210R CPUs @ 2.40GHz, and the Ubuntu 20.04.6 LTS operating system. The initial learning rate is set to 0.001, decaying linearly to zero over the training process. The batch size is set to 16, and the training spans 200 epochs. The input resolution for images during training is 1024 × 1024. For the VisDrone dataset, experiments are carried out using the original training and validation sets. For the UAVDT dataset, we manually screened and deleted 5 video clips that lacked a large number of annotations, and divided the remaining 45 video sequences for multi-target detection and tracking into training and validation sets in a 4:1 ratio. To reduce the computation of redundant images, one frame is extracted every ten frames for continuous video frames. In order to improve the generalization ability of the model, data augmentation techniques such as random flipping and random color jitter are applied to the input images.

\subsection{Evaluation Metrics}
In our experiments, we employ widely used evaluation metrics, including Average Precision (AP) and mean Average Precision (mAP), to assess the accuracy of our model. Precision reflects the proportion of true positives (TP) among all predicted positives, which includes both true positives and false positives (FP). Recall indicates the proportion of true positives among all actual positive outcomes, which includes both true positives and false negatives (FN). The formulas for Precision (P) and Recall (R) are as follows:

\begin{equation}
	\begin{split}
		P=\frac{\mathrm{TP}}{\mathrm{TP}+\mathrm{FP}},   \label{Eq:19}
	\end{split}
\end{equation}

\begin{equation}
	\begin{split}
		R=\frac{\mathrm{TP}}{\mathrm{TP}+\mathrm{FN}},   \label{Eq:20}
	\end{split}
\end{equation}

AP and mAP are defined as follows:

\begin{equation}
	\begin{split}
		\text{AP}=\int_0^1P(R)dR,   \label{Eq:21}
	\end{split}
\end{equation}

\begin{equation}
	\begin{split}
		\text{mAP}=\frac1{N_{\mathrm{cls}}}\sum_{i=1}^{N_{\mathrm{cls}}}\mathrm{AP}_{i}, \label{Eq:22}
	\end{split}
\end{equation}
where $N_{\mathrm{cls}}$ represents the total number of categories.

\subsection{Comparative Experiment}

In this section, we compare the DREB-Net with widely used methods in object detection. These include one-stage and two-stage detection algorithms, anchor-based and anchor-free approaches, methods based on Convolutional Neural Networks (CNNs) and attention mechanisms, as well as transformer-based methods. The specific algorithms compared include Faster R-CNN~\cite{ren2015faster}, RetinaNet~\cite{lin2017focal}, CornerNet~\cite{law2018cornernet}, CenterNet~\cite{zhou2019objects}, FCOS~\cite{tian2022fully}, FSAF~\cite{zhu2019feature}, YOLOX~\cite{ge2021yolox}, YOLOF~\cite{chen2021you}, VFNet~\cite{zhang2021varifocalnet}, CentripetalNet~\cite{dong2020centripetalnet} and DINO~\cite{zhang2022dino}. For a fair comparison, we retrain all methods on the blurred dataset.

1) Test results on the VisDrone-2019-DET dataset: We utilized the official training and testing sets, which contained 6471 and 548 images, respectively. The official dataset provided annotations for 11 categories, including pedestrian, people, bicycle, car, van, truck, tricycle, awning-tricycle, bus, motor, and others. This dataset labels dense small targets that are difficult to label as ignored regions. During the training process, we filled these regions with black. For this dataset, due to the complexity of the scenes and the cost of annotation, we merged similar categories, such as pedestrians and people. Moreover, we removed non-interest categories, ultimately retaining four classes: people, car, truck, and bus. We compared our method with the algorithms above and reported detection accuracy and recall rates for each category and small, medium, and large targets across the dataset.

\begin{figure}[tb!]
	\centering
	\includegraphics[width=85mm]{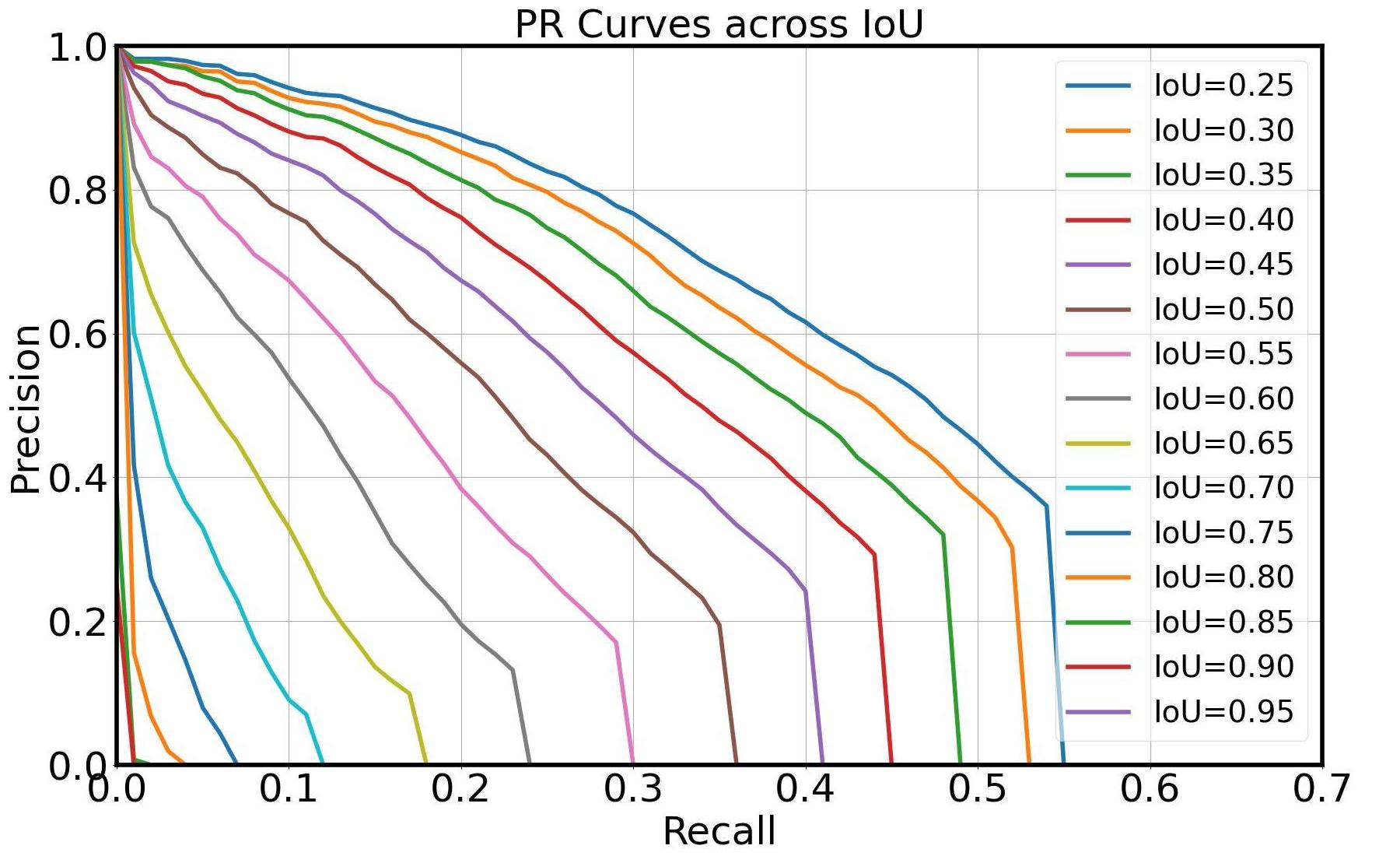}
	\caption{PR Curves of DREB-Net at Different IoU Thresholds on the blurred VisDrone Dataset.}
	\label{fig:Curve_IoU}
\end{figure}

\begin{figure}[tb!]
	\centering
	\includegraphics[width=85mm]{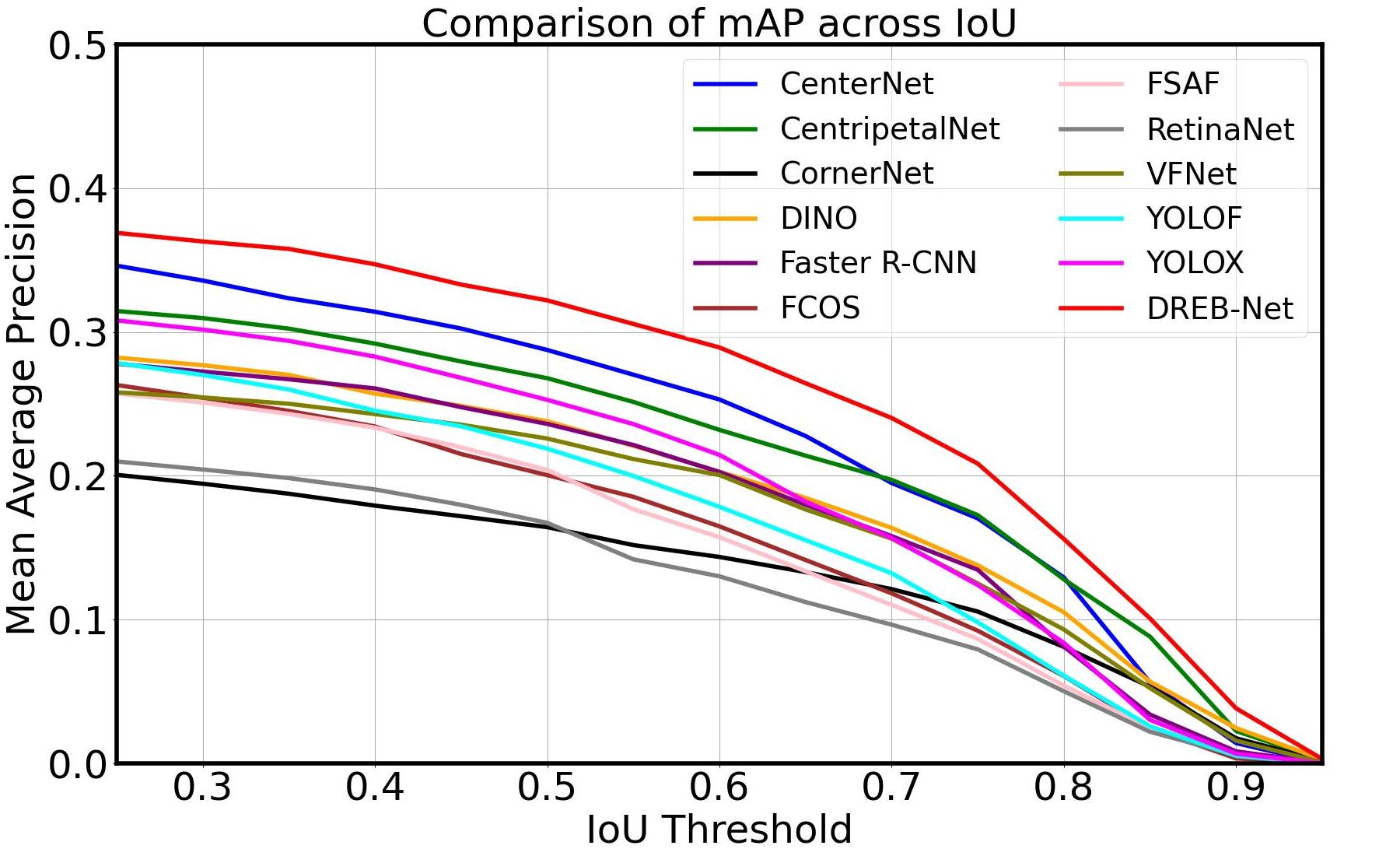}
	\caption{Comparison of mAP Curves across IoU Thresholds on the blurred VisDrone Dataset.}
	\label{fig:mAP_comparison}
\end{figure}

\begin{figure*}[htbp]
	\centering
	\begin{subfigure}{0.48\textwidth}
		\includegraphics[width=\linewidth]{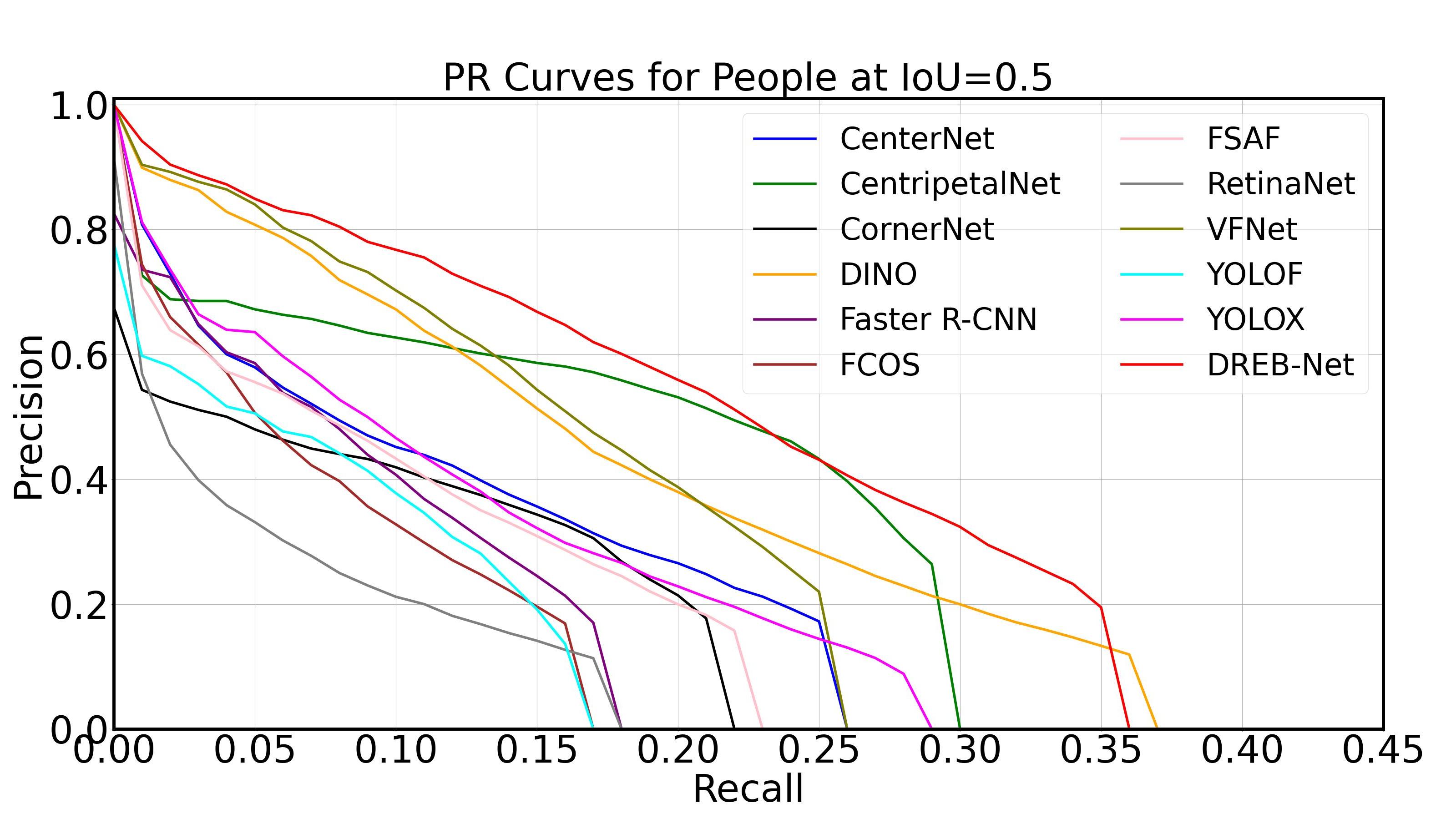}
		\caption*{(a)}
	\end{subfigure}
	\hfill
	\begin{subfigure}{0.48\textwidth}
		\includegraphics[width=\linewidth]{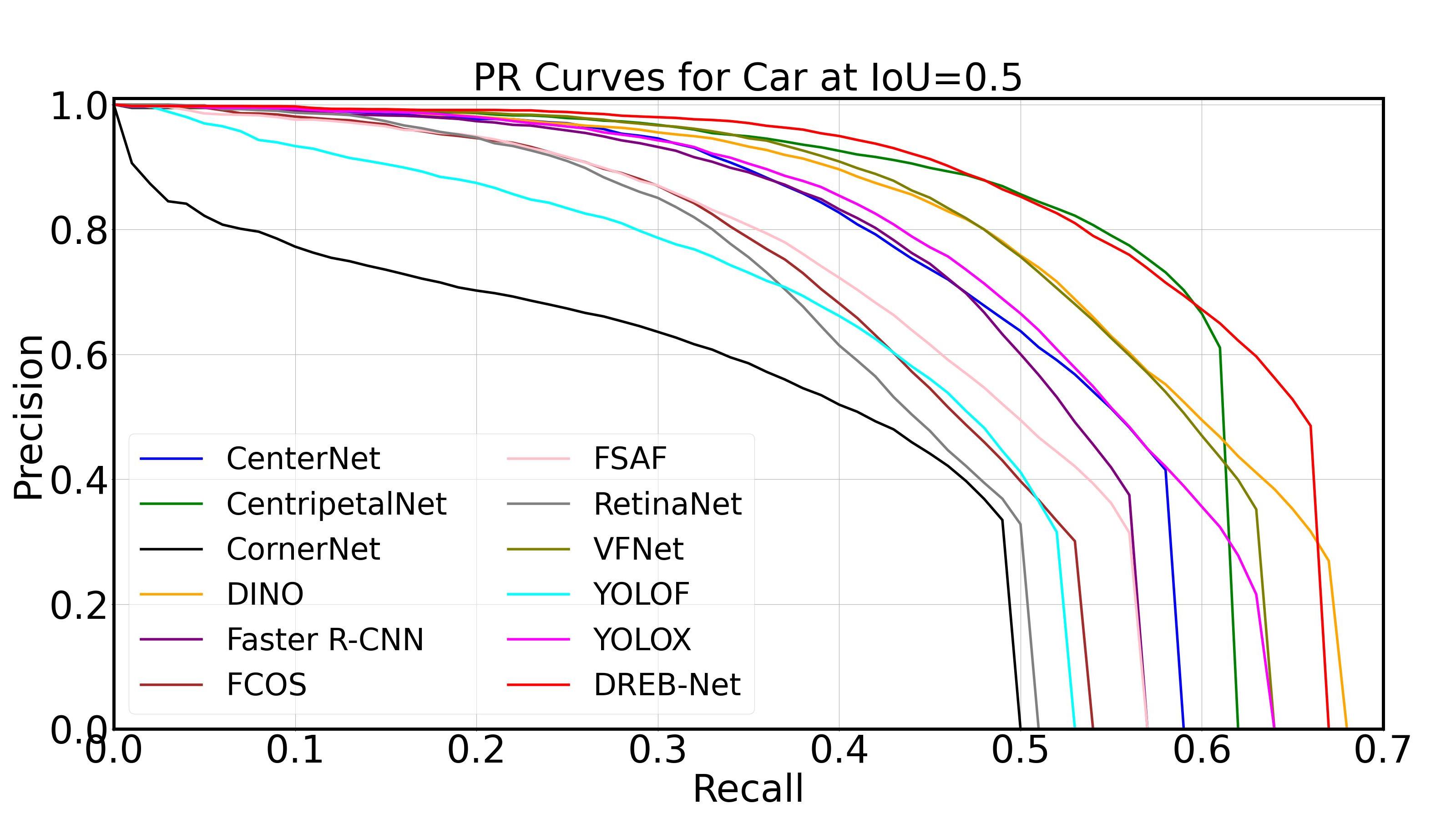}
		\caption*{(b)}
	\end{subfigure}

	\begin{subfigure}{0.48\textwidth}
		\includegraphics[width=\linewidth]{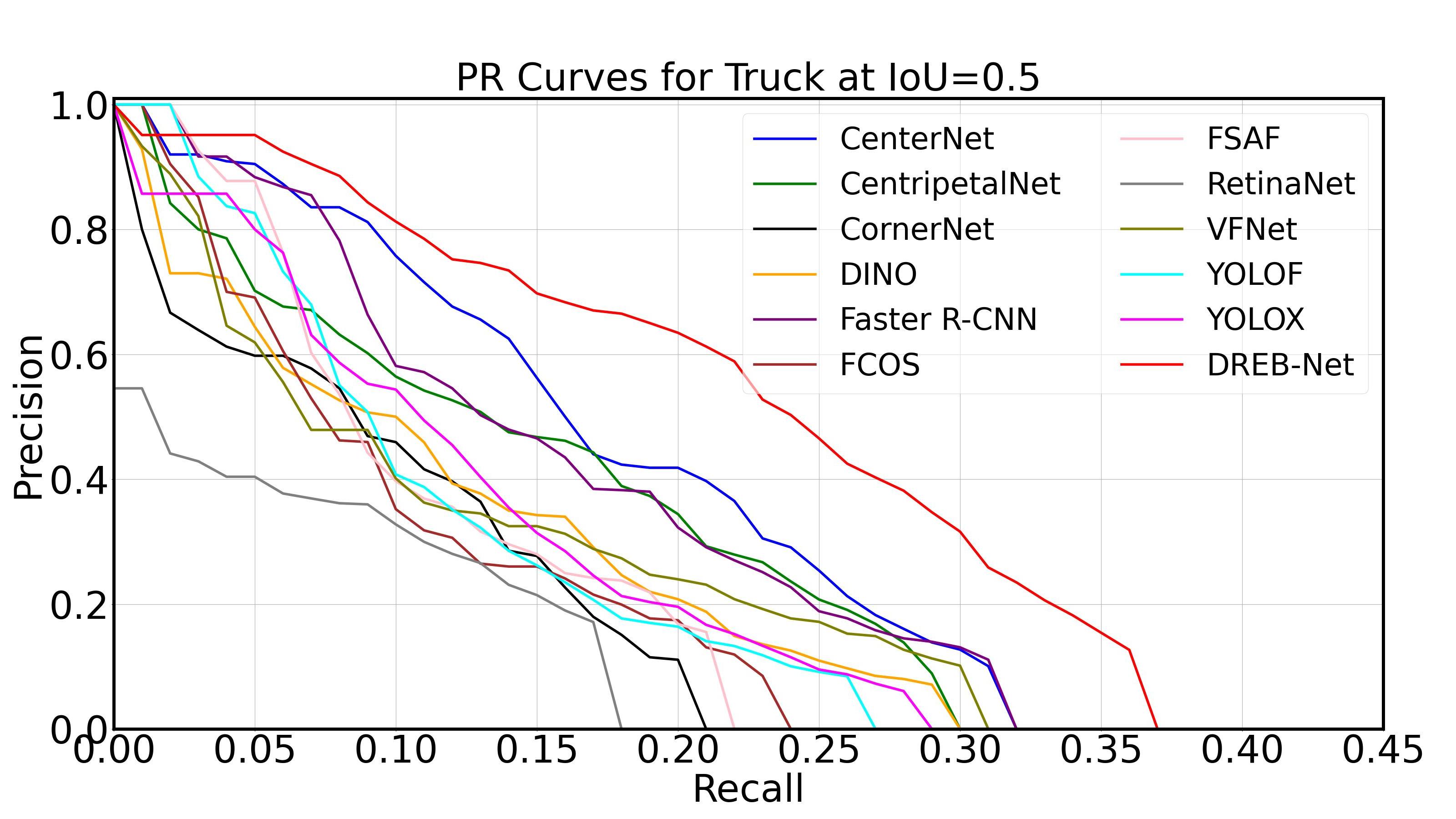}
		\caption*{(c)}
	\end{subfigure}
	\hfill
	\begin{subfigure}{0.48\textwidth}
		\includegraphics[width=\linewidth]{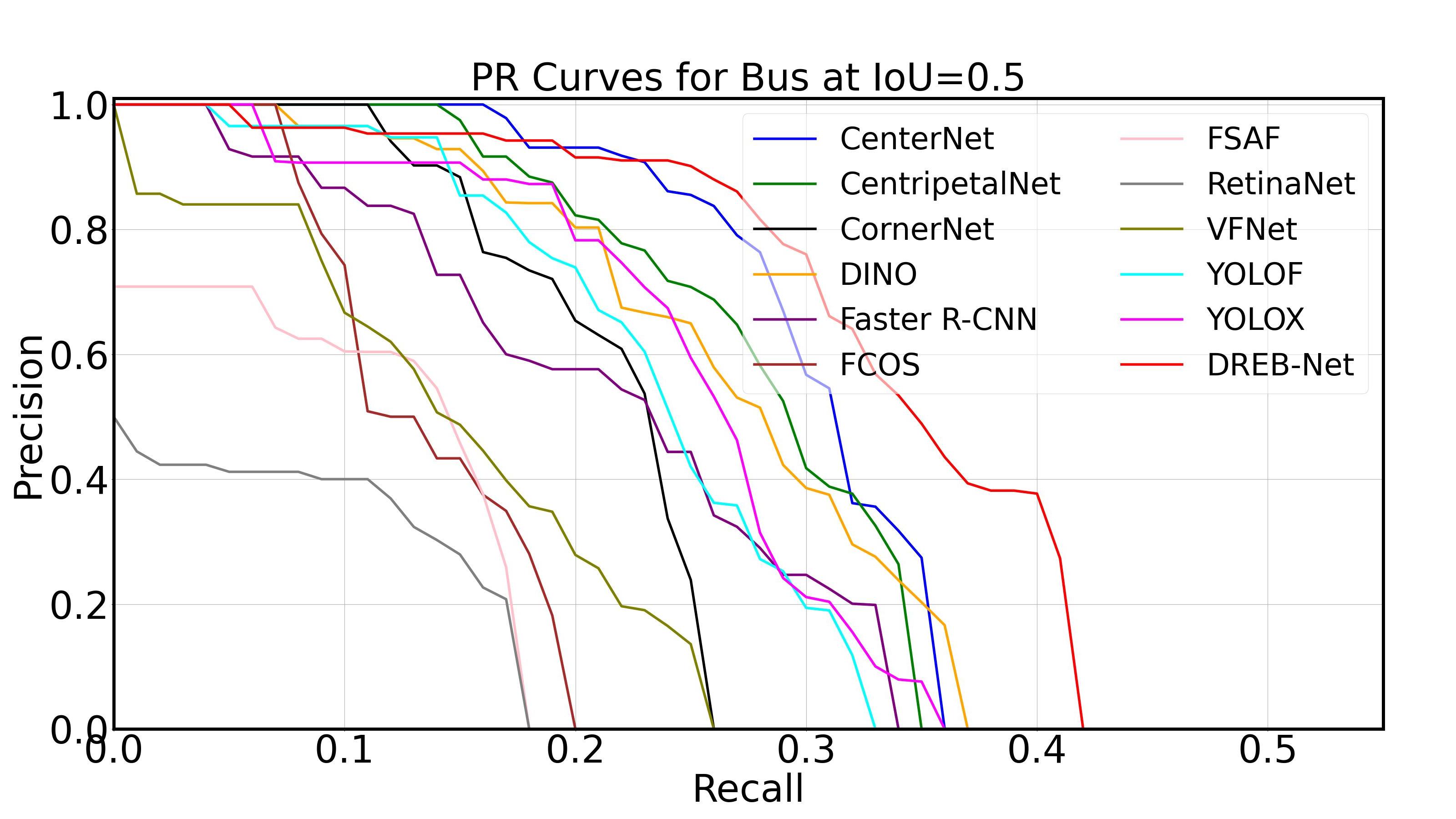}
		\caption*{(d)}
	\end{subfigure}
	\caption{Category-wise Precision-Recall Curves at IoU=0.5 on the blurred VisDrone dataset. (a) PR Curves for people. (b) PR Curves for car. (c) PR Curves for truck. (d) PR Curves for bus.}
	\label{fig:curves_PR}
\end{figure*}

Table~\ref{table1} shows the comparative results of our model with existing methods on the VisDrone dataset. Experiments show that our method has significant advantages in both detection accuracy and recall rates, especially in small-sized objects such as people. Specifically, detection accuracies for the four categories reached 0.213, 0.599, 0.226, and 0.337, with recall rates of 0.388, 0.430, 0.342, and 0.469, respectively. Accuracies for small, medium, and large targets were 0.161, 0.523, and 0.792 respectively, while the overall dataset's average accuracy and recall were 0.344 and 0.448. Compared to the best-performing existing model, our method achieved the best results in the accuracy of small, medium, and large targets, as well as the accuracy and recall of the entire dataset. These results clearly demonstrate the superiority of our method, which can be efficiently applied to object detection tasks in drone blurry images. In Figure~\ref{fig5}, we visualize the detection results of the proposed method. It can be seen that despite occasional missed or false detections caused by excessive blurring, DREB-Net still exhibits good performance on blurred images of drones.

Furthermore, to provide a more comprehensive explanation of our method, we provide a richer set of curve graphs:

\textbf{Precision-Recall (PR) Curves at Different IoU Thresholds}: As shown in Figure~\ref{fig:Curve_IoU}, we present the precision-recall curves of the DREB-Net under various Intersection over Union (IoU) threshold conditions. Observing the change from IoU=0.30 to IoU=0.95, we note that as the IoU threshold increases, the precision gradually decreases under higher recall conditions. This indicates that higher IoU thresholds pose a greater challenge to model performance and demand more precise localization capabilities.

\textbf{Comparison Curves of mAP across IoU.}: In Figure~\ref{fig:mAP_comparison}, we report the comparison curves of mean Average Precision (mAP) across multiple methods under various Intersection over Union (IoU) thresholds. The results indicate that as the IoU threshold increases, the mAP for all methods tends to decrease. However, across all tested IoU thresholds, our proposed DREB-Net model consistently demonstrates the highest mAP performance.

\textbf{Category-wise Precision-Recall Curves at IoU=0.5}: Figure~\ref{fig:curves_PR} presents the precision-recall curve for different categories (people, car, truck, bus) under the condition of IoU=0.5. The results show that at this threshold, the detection precision for cars is higher than that of other categories, while the precision and recall for trucks are relatively lower. This may be related to the fewer training samples for trucks and the difficulty of recognizing in blurry images.

\begin{figure*}[htbp]
	\centering
	\begin{subfigure}{0.48\textwidth}
		\includegraphics[width=\linewidth]{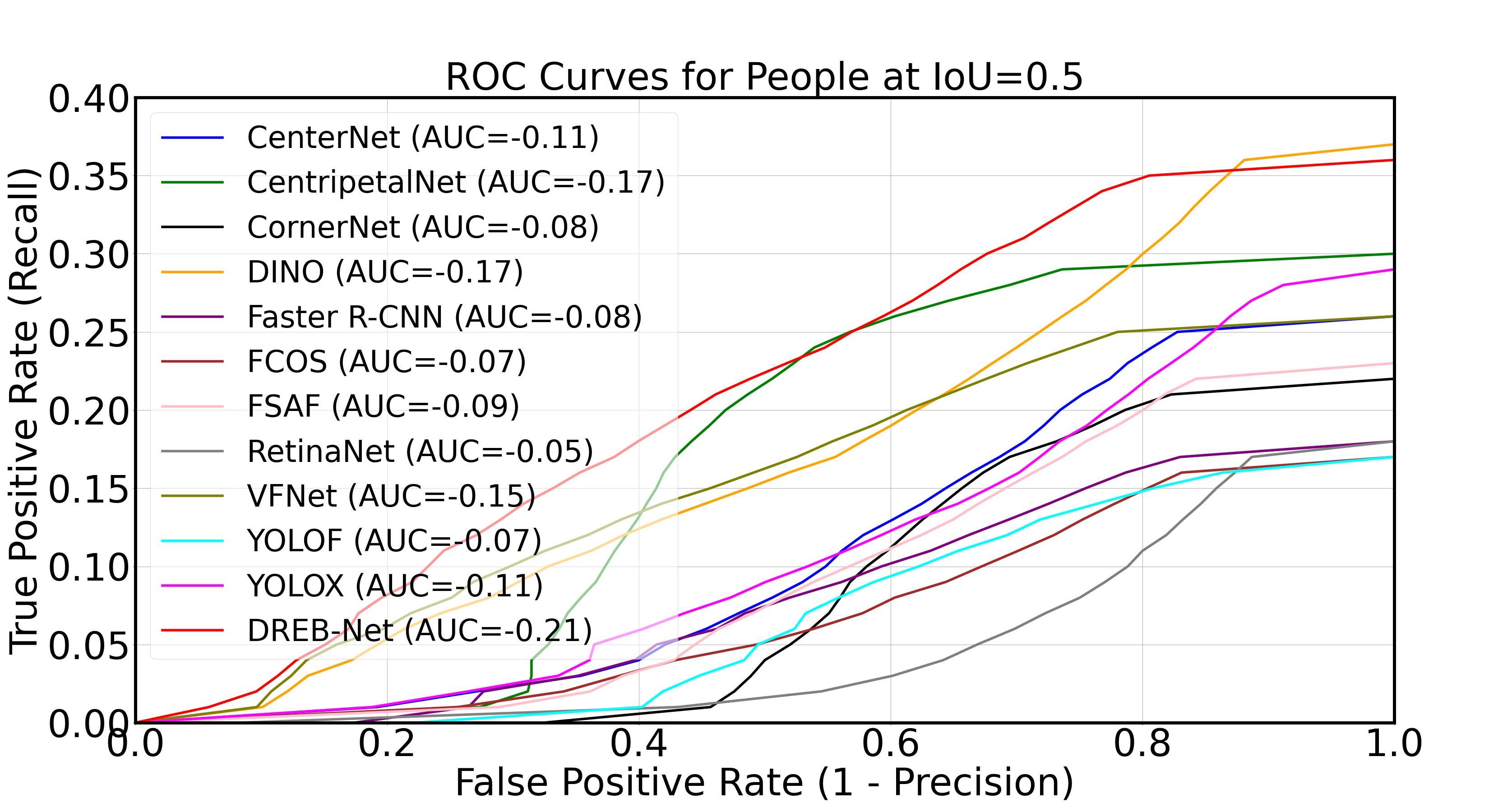}
		\caption*{(a)}
	\end{subfigure}
	\hfill
	\begin{subfigure}{0.48\textwidth}
		\includegraphics[width=\linewidth]{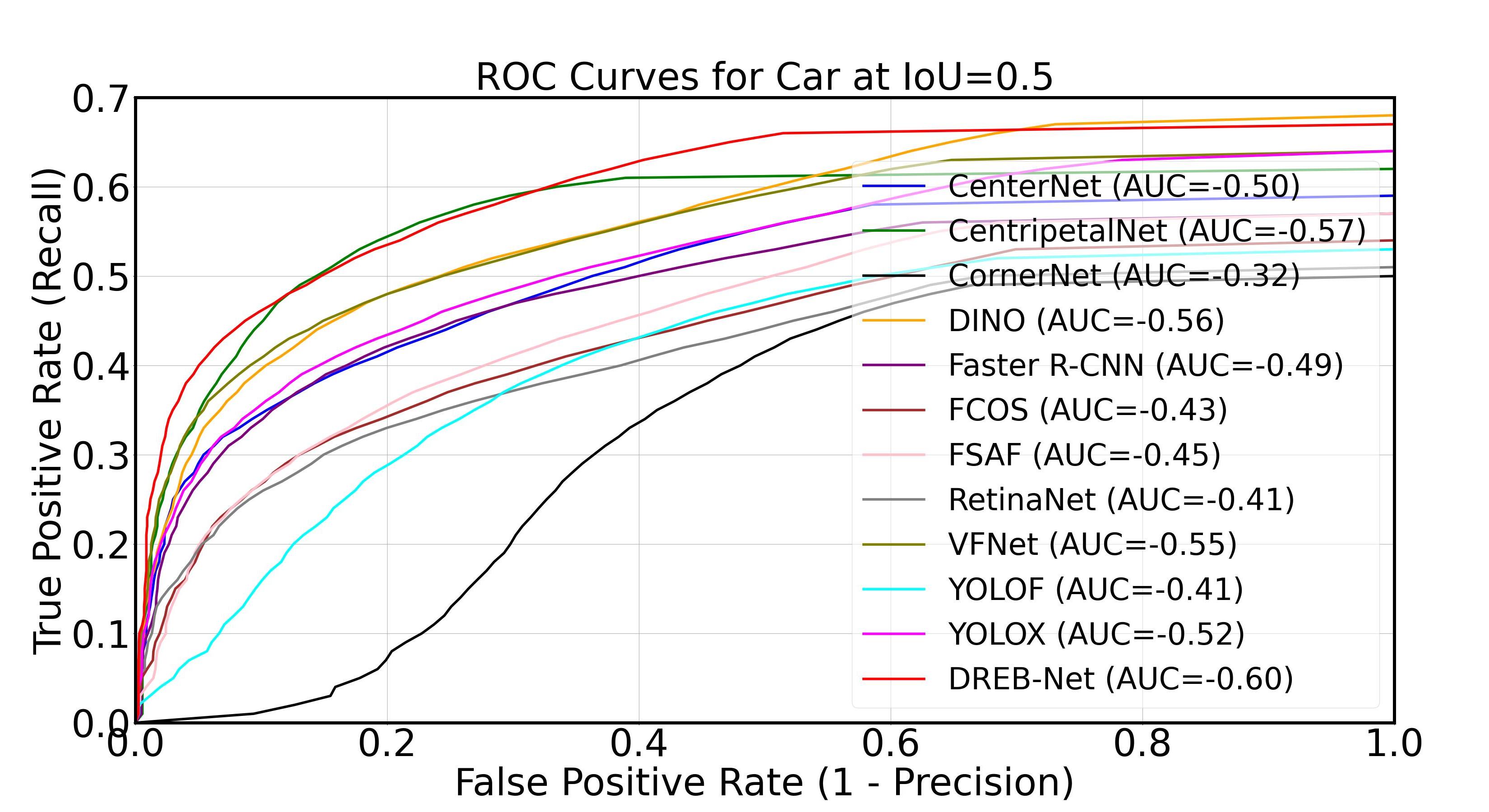}
		\caption*{(b)}
	\end{subfigure}
	
	\begin{subfigure}{0.48\textwidth}
		\includegraphics[width=\linewidth]{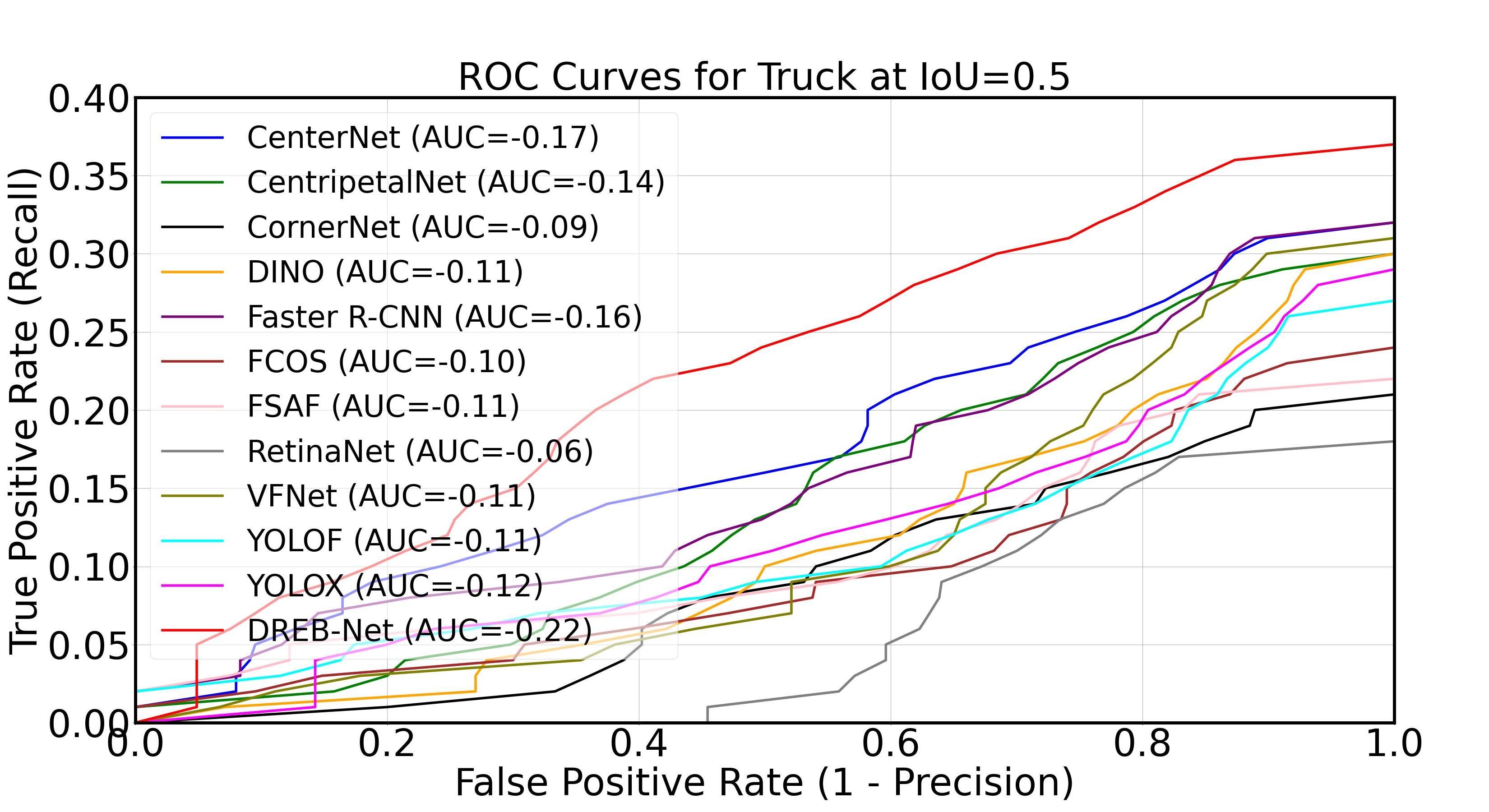}
		\caption*{(c)}
	\end{subfigure}
	\hfill
	\begin{subfigure}{0.48\textwidth}
		\includegraphics[width=\linewidth]{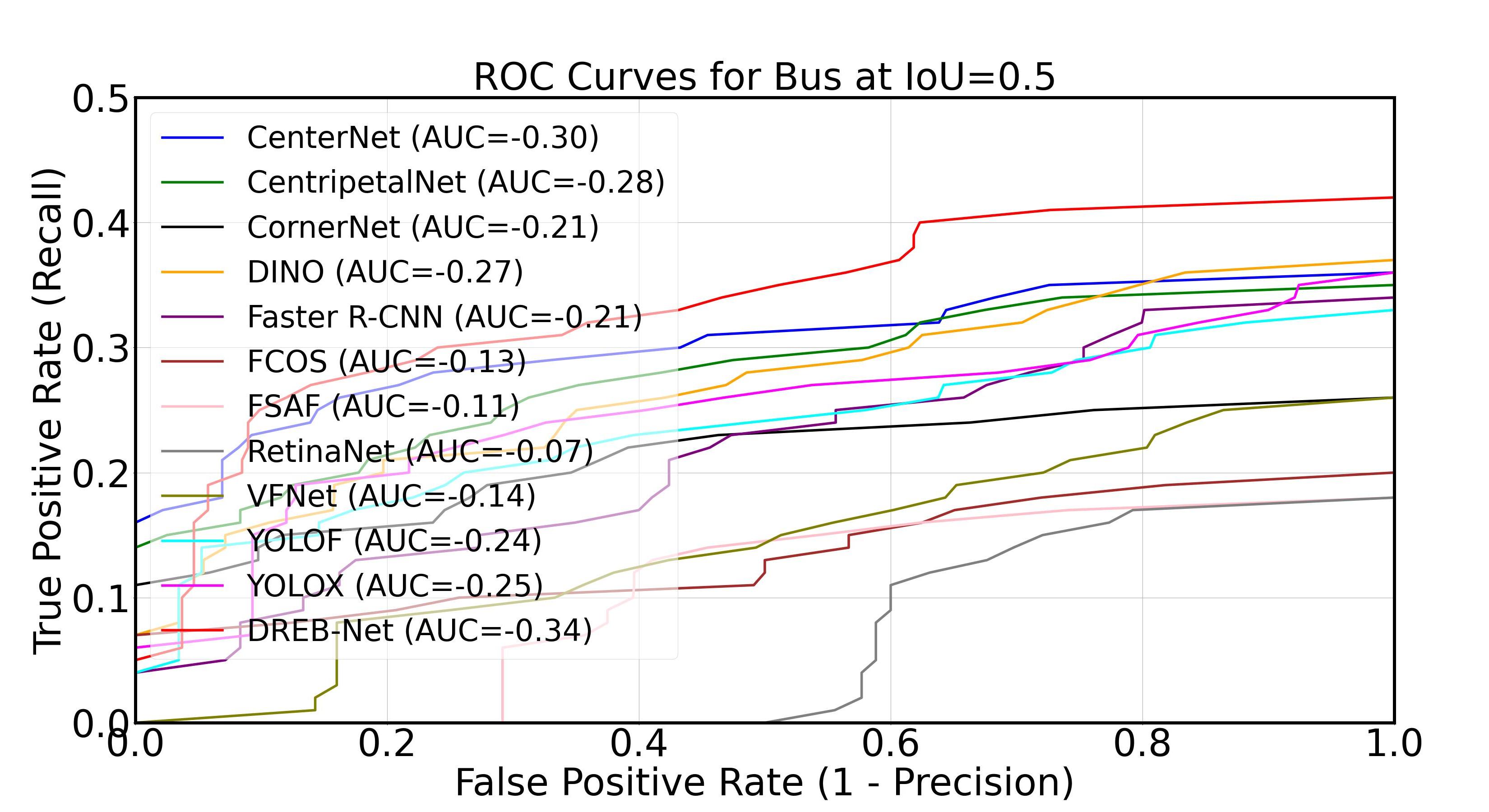}
		\caption*{(d)}
	\end{subfigure}
	\caption{ROC Curves for Each Category at IoU=0.5 on the blurred VisDrone Dataset. (a) ROC Curves for people. (b) ROC Curves for car. (c) ROC Curves for truck. (d) ROC Curves for bus.}
	\label{fig:ROC_curves}
\end{figure*}

\textbf{Category-wise Receiver Operating Characteristic (ROC) Curves at IoU=0.5}: Figure~\ref{fig:ROC_curves} shows the ROC curves and corresponding AUC values for each category. As can be seen from the figure, the car category has the highest AUC value (0.60), indicating that it has better classification performance across various thresholds. In contrast, the AUC values of truck and bus are lower, which may be due to the greater difficulty in accurately recognizing these categories in blurry images.

2) Test results on the UAVDT dataset: Although the UAVDT dataset is widely used in drone-based object detection tasks, the official dataset was annotated for multi-target tracking, without specific training and testing sets for detection tasks. We randomly divided the video clips in a 4:1 ratio and constructed the object detection dataset at three frames per second. The divided dataset contains 3050 training images and 736 testing images. Due to the varying lengths of each video segment, the training-testing ratio for images did not correspond exactly to that of the video segments. We used the original annotations provided by the official, including three types of objects: car, truck, and bus. The areas marked as ignored regions were filled with black during the training phase.

\begin{table*}[htbp!]
	\centering
	\caption{Test Results on The Blurred UAVDT Dataset.}
	\label{table2}
	\begin{tabular}{ccccccccccccc}
		\hline
		\multirow{2}{*}{{Models}} & \multicolumn{3}{c}{$AP_{50}$} & \multicolumn{3}{c}{$AR_{50}$} & \multirow{2}{*}{$mAP_{s}$} & \multirow{2}{*}{$mAP_{m}$} & \multirow{2}{*}{$mAP_{l}$} & \multirow{2}{*}{$mAP_{50}$} & \multirow{2}{*}{$mAR_{50}$}   & \multirow{2}{*}{FPS} \\ \cline{2-7}
		& {car} & {truck} & {bus} & {car} & {truck} & {bus} & & & & & & \\ \hline
		{Faster R-CNN~\cite{ren2015faster}} & 0.453 & 0.047 & 0.061 & 0.434 & 0.309 & 0.255 & 0.063 & 0.283 & 0.409 & 0.187 & 0.340   & 16.52\\ \hline
		{RetinaNet~\cite{lin2017focal}} & 0.427 & 0.008 & 0.047 & 0.447 & 0.296 & \textbf{0.636} & 0.050 & 0.260 & 0.350 & 0.161 & 0.488    & 17.29\\ \hline
		{CornerNet~\cite{law2018cornernet}} & 0.439 & 0.032 & 0.105 & 0.440 & 0.294 & 0.399 & 0.111 & 0.322 & 0.182 & 0.192 & 0.397   & 2.99\\ \hline
		{CenterNet~\cite{zhou2019objects}} & 0.475 & 0.036 & 0.117 & 0.430 & 0.347 & 0.315 & 0.083 & 0.282 & 0.479 & 0.209 & 0.433   & 27.02 \\ \hline
		{FCOS~\cite{tian2022fully}} & 0.422 & \textbf{0.104} & 0.102 & 0.448 & \textbf{0.468} & 0.621 & 0.056 & 0.328 & 0.369 & 0.209 & 0.494 &17.56 \\ \hline
		{FSAF~\cite{zhu2019feature}} & 0.441 & 0.02 & 0.013 & 0.446 & 0.312 & 0.216 & 0.058 & 0.255 & 0.389 & 0.158 & 0.369 & 17.72  \\ \hline
		{YOLOX~\cite{ge2021yolox}} & 0.481 & 0.026 & 0.034 & 0.451 & 0.096 & 0.117 & 0.093 & 0.252 & 0.316 & 0.180 & 0.279 & 37.88  \\ \hline
		{YOLOF~\cite{chen2021you}} & 0.336 & 0.022 & 0.087 & 0.435 & 0.224 & 0.36 & 0.103 & 0.243 & 0.362 & 0.148 & 0.362 & 29.63 \\ \hline
		{VFNet~\cite{zhang2021varifocalnet}} & 0.479 & 0.005 & 0.031 & 0.463 & 0.155 & 0.420 & 0.076 & 0.259 & 0.323 & 0.171 & 0.433 & 14.77 \\ \hline
		{CentripetalNet~\cite{dong2020centripetalnet}} & 0.517 & 0.037 & 0.086 & 0.450 & 0.392 & 0.405 & 0.109 & 0.311 & 0.389 & 0.213 & 0.442 & 2.43  \\ \hline
		{DINO~\cite{zhang2022dino}} & 0.435 & 0.025 & 0.103 & 0.440 & 0.347 & 0.584 & 0.067 & 0.283 & 0.474 & 0.188 & 0.497 & 8.02 \\ \hline
		{DREB-Net\_tiny} & 0.537 & 0.025 & 0.170 & 0.457 & 0.362 & 0.459 & 0.101 & 0.338 & 0.521 & 0.244 & 0.498 & 32.08 \\ \hline
		\textbf{DREB-Net} & \textbf{0.566} & 0.038 & \textbf{0.223} & \textbf{0.470} & 0.372 & 0.459 & \textbf{0.114} & \textbf{0.353} & \textbf{0.558} & \textbf{0.276} & \textbf{0.502} & 11.73 \\ \hline		
	\end{tabular}
\end{table*}

\begin{table*}[htbp!]
	\centering
	\caption{Ablation Experiment on The Blurred VisDrone-2019-DET Dataset.}
	\label{table3}
	\setlength{\tabcolsep}{4pt} 
	\begin{tabular}{cccccccccccccccc}
		\hline
		\multirow{2}{*}{{BRAB}} & \multirow{2}{*}{{MAGFF}} & \multirow{2}{*}{{LFAMM}} & \multicolumn{4}{c}{$AP_{50}$} & \multicolumn{4}{c}{$AR_{50}$} & \multirow{2}{*}{$mAP_{s}$} & \multirow{2}{*}{$mAP_{m}$} & \multirow{2}{*}{$mAP_{l}$} & \multirow{2}{*}{$mAP_{50}$} & \multirow{2}{*}{$mAR_{50}$} \\ \cline{4-11}
		& & & {people} & {car} & {truck} & {bus} & {people} & {car} & {truck} & {bus} & & & & &  \\ \hline
		{$\times$} & {$\times$} & {$\times$} & 0.118 & 0.514 & 0.186 & 0.314 & 0.312 & 0.409 & 0.339 & 0.456 & 0.095 & 0.457 & 0.742 & 0.283 & 0.386   \\ \hline
		{\checkmark} & {$\times$} & {$\times$} & 0.153 & 0.562 & 0.203 & 0.298 & 0.346 & 0.421 & 0.347 & 0.453 & 0.117 & 0.479 & 0.773 & 0.304 & 0.405   \\ \hline
		{$\times$} & {$\times$} & {\checkmark} & 0.147 & 0.560 & 0.188 & 0.320 & 0.338 & 0.422 & 0.350 & 0.469 & 0.114 & 0.487 & 0.759 & 0.304 & 0.408   \\ \hline
		{\checkmark} & {\checkmark} & {$\times$} & 0.196 & 0.591 & 0.218 & 0.331 & 0.373 & 0.427 & 0.338 & 0.469 & 0.146 & 0.508 & 0.788 & 0.334 & 0.437   \\ \hline
		{\checkmark} & {$\times$} & {\checkmark} & 0.174 & 0.583 & 0.208 & 0.325 & 0.357 & 0.425 & 0.365 & 0.460 & 0.134 & 0.508 & 0.758 & 0.323 & 0.428   \\ \hline
		{\checkmark} & {\checkmark} & {\checkmark} & \textbf{0.213} & \textbf{0.599} & \textbf{0.226} & \textbf{0.337} & \textbf{0.388} & \textbf{0.430} & \textbf{0.342} & \textbf{0.469} & \textbf{0.161} & \textbf{0.523} & \textbf{0.792} & \textbf{0.344} & \textbf{0.448}   \\ \hline
	\end{tabular}
\end{table*}

\begin{table}[htbp!]
	\centering
	\caption{FLOPs and parameters of DREB-Net and comparison with CenterNet-res50}
	\label{table4}
	\setlength{\tabcolsep}{2pt} 
	\begin{tabular}{cccccc}
		\hline
		Model & BRAB & MAGFF & LFAMM &  FLOPs(G) & Para.(val/total, MB) \\ \hline
		{CenterNet} & {$\times$} & {$\times$} & {$\times$} & 235.75 &  34.44 \\ \hline
		{DREB-Net} & {$\times$} & {$\times$} & {$\times$} & 193.24 &   17.93  \\ \hline
		{DREB-Net} & {\checkmark} & {$\times$} & {$\times$} & 206.60 & 17.87 / 29.18 \\ \hline
		{DREB-Net} & {$\times$} & {$\times$} & {\checkmark} & 193.24 & 18.46 \\ \hline
		{DREB-Net} & {\checkmark} & {\checkmark} & {$\times$} & 206.90 & 17.91 / 29.22 \\ \hline
		{DREB-Net} & {\checkmark} & {$\times$} & {\checkmark} & 206.60 & 18.94 / 30.25  \\ \hline
		{DREB-Net} & {\checkmark} & {\checkmark} & {\checkmark} & 206.90 & 18.97 / 30.28 \\ \hline
		
	\end{tabular}
\end{table}

\begin{figure}[tb!]
	\centering
	\includegraphics[width=85mm]{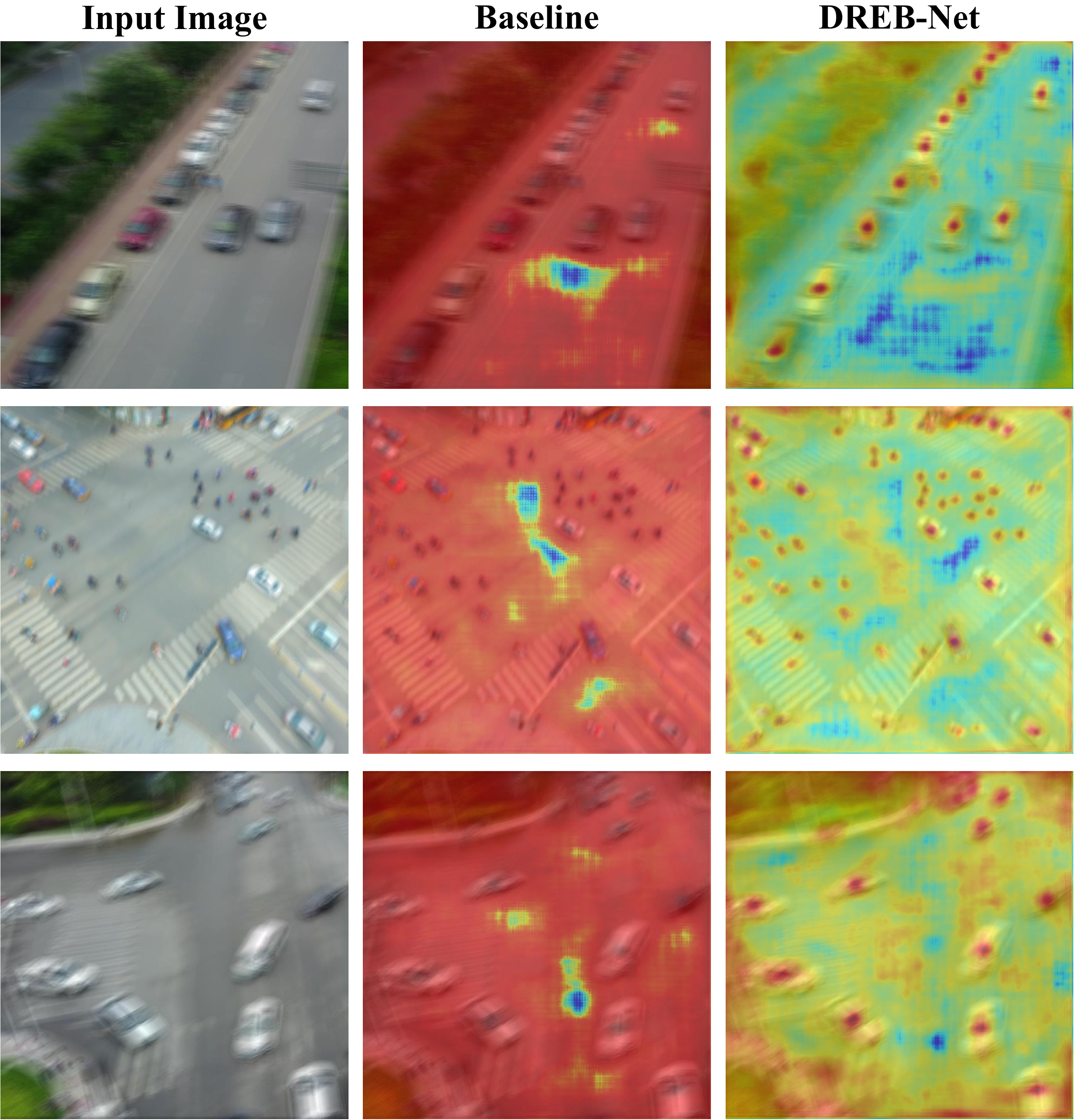}
	\caption{Heatmaps Comparison of Object Detection Generated Using the Baseline Method and Our proposed DREB-Net.}
	\label{heatmap}
\end{figure}

Table~\ref{table2} shows the comparison of DREB-Net with existing methods on the UAVDT dataset. Although the UAVDT dataset contains a higher proportion of night-time and high-altitude images, making the scenes relatively more complex, and consists only of several dozen video segments, which limits its richness, the overall performance of DREB-Net is not as strong as on the VisDrone dataset. However, it still demonstrates advantages compared to other existing methods. The accuracies for the categories of car, truck, and bus were 0.566, 0.038, and 0.223, with recall rates of 0.470, 0.372, and 0.459, respectively. For small, medium, and large objects, the accuracy reached 0.114, 0.353, and 0.558, respectively, and the overall average accuracy and recall of the dataset were 0.276 and 0.502. Compared to the best-performing model, our method achieved the best results in the accuracy of small, medium, and large targets, as well as the accuracy and recall of the entire dataset. Figure~\ref{fig6} shows the object detection performance of our method on the UAVDT dataset, affirming its applicability and advantages in complex and blurry drone image environments.

\begin{figure*}[htbp!]
	\centering
	\includegraphics[width=180mm]{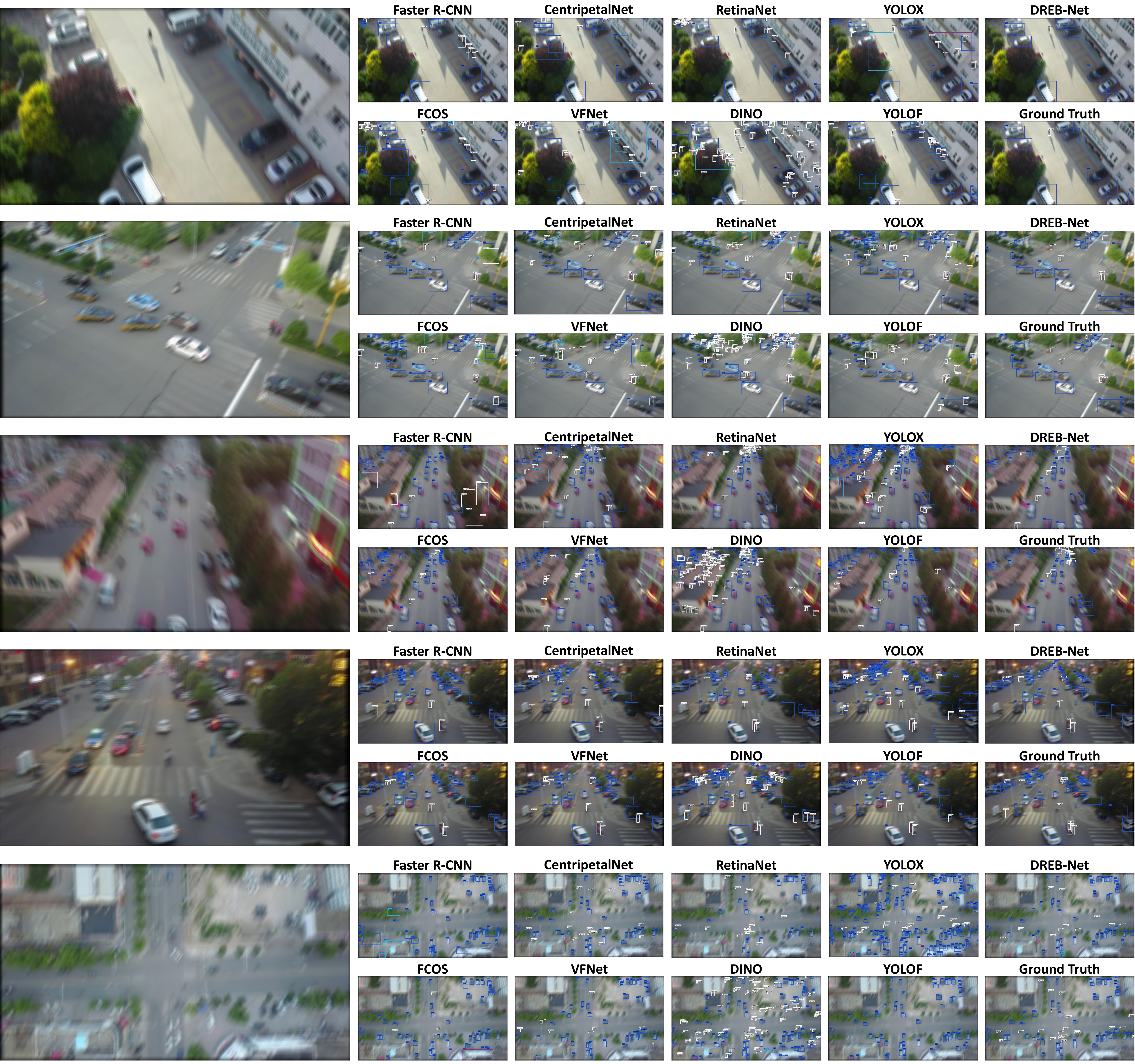}
	\caption{Visualization of the detection results of DREB-Net tested on the blurred VisDrone-2019-DET dataset. To better demonstrate the effect, we provide a comparison of the detection results of our DREB-Net, existing object detection methods, and Ground Truth.}
	\label{fig5}
\end{figure*}

\begin{figure*}[htbp!]
	\centering
	\includegraphics[width=180mm]{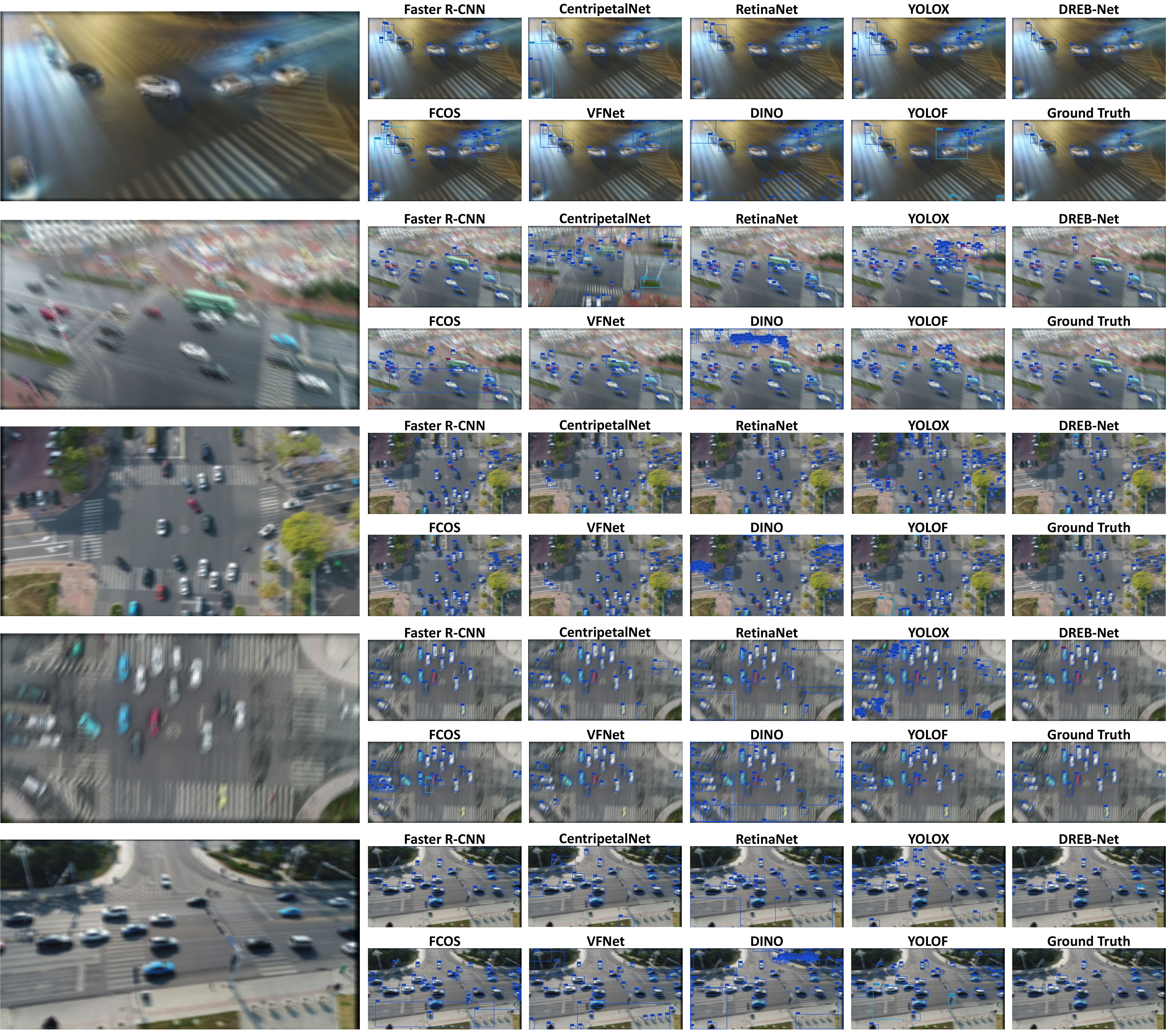}
	\caption{Visualization of the detection results of DREB-Net tested on the blurred UAVDT dataset. As in the previous figure, we provide a comparison of the detection results of our DREB-Net, existing object detection methods, and Ground Truth.}
	\label{fig6}
\end{figure*}

\begin{figure*}[htbp!]
	\centering
	\includegraphics[width=180mm]{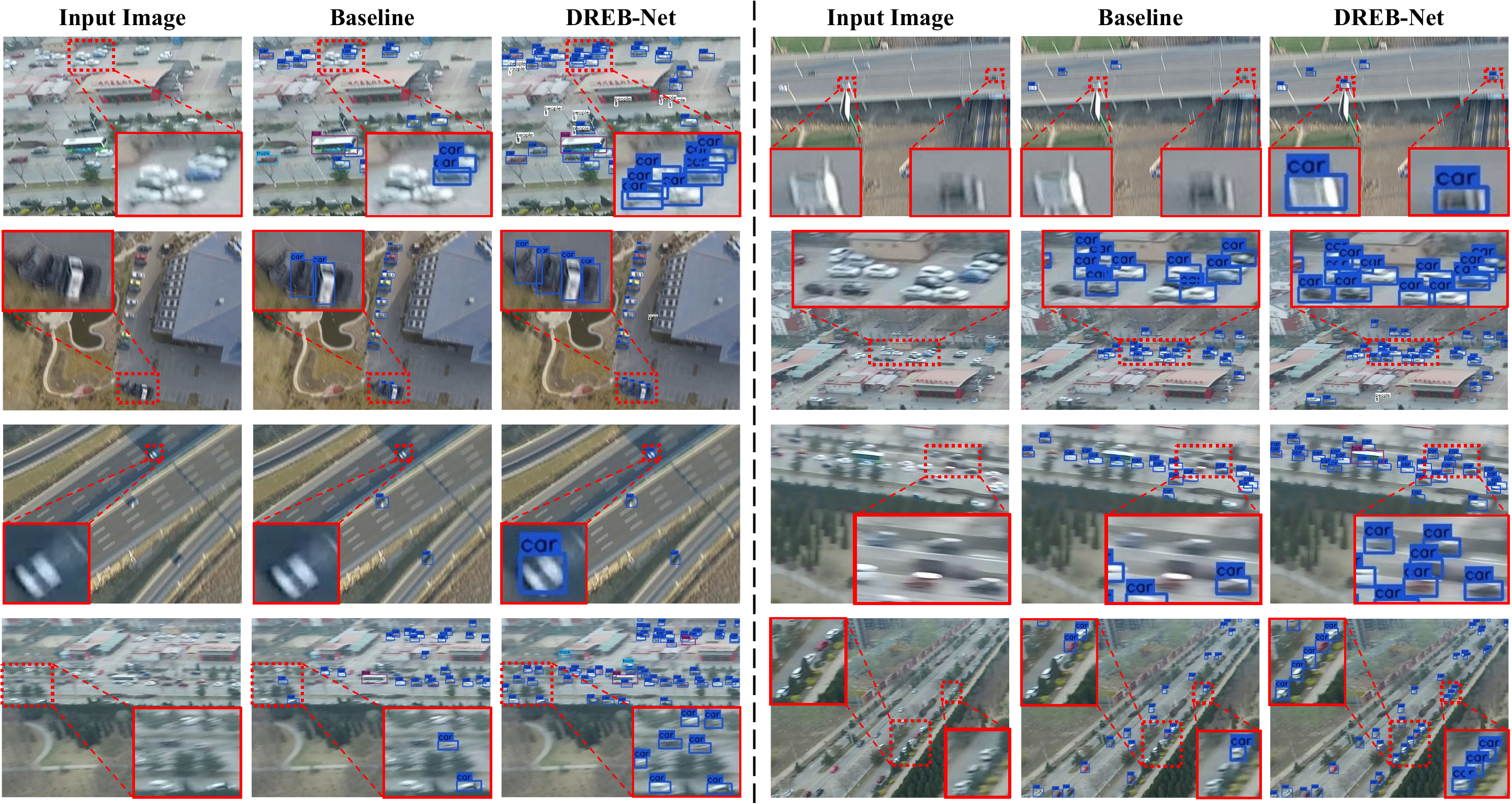}
	\caption{Visualization of the detection results of DREB-Net (pretrained on the VisDrone-2019-DET dataset, without any other training) tested on real blurry images.}
	\label{real_blur}
\end{figure*}

\subsection{Ablation Experiments}
In order to comprehensively evaluate the DREB-Net, we conducted ablation experiments on the VisDrone dataset. These experiments aim to analyze the impact of BRAB, MAGFF, and LFAMM on overall performance. The experimental results are shown in Table~\ref{table3}. The specific analysis of the ablation experiment results is as follows:

1) BRAB: The BRAB is designed to enhance the model's ability to detect objects in blurred images by optimizing the clarity of shallow features. As shown in Table 1, compared with the baseline model by adding only the BRAB module, the $AP_{50}$ of small, medium, and large objects increased by 0.022, 0.022, and 0.031 respectively, while the overall $mAP_{50}$ and $mAR_{50}$ increased by 0.021 and 0.019 respectively. Despite a slight decline in the metrics for buses, there were significant improvements in accuracy and recall for people, cars, and trucks. When BRAB and FLAMM are used simultaneously, compared to using FLAMM alone, it significantly enhances detection performance for small and medium objects, with improvements of 0.020 and 0.021 in $mAP_{s}$ and $mAP_{m}$ respectively, and increases of 0.027, 0.023, 0.020, and 0.005 in the four categories respectively. The overall $mAP_{50}$ and $mAR_{50}$ increased by 0.019 and 0.020, respectively. It indicates that the BRAB is effective in enhancing the recognition ability of the model under blurry conditions, especially for smaller and more detailed targets.

2) MAGFF: The MAGFF module, through an innovative feature fusion strategy, optimizes the information flow between features. Since MAGFF is a fusion module between the BRAB and the object detection main branch, it can only be evaluated in configurations where BRAB is used. Under this condition, we compared the model's performance changes with the addition of MAGFF configuration. From using BRAB alone to introducing MAGFF into the model, the detection performance of the model has significantly improved in all four categories, with increases of 0.043, 0.029, 0.015, and 0.033 in $AP_{50}$, and 0.029, 0.029, and 0.015 in $mAP_{s}$, $mAP_{m}$, $mAP_{l}$ respectively, and 0.03 and 0.032 in $mAP_{50}$ and $mAR_{50}$ metrics for all categories respectively. Under the condition of simultaneously using BRAB and LFAMM, the addition of the MAGFF module showed significant improvement in detection accuracy for all four categories, with increments of 0.039, 0.016, 0.018, and 0.031 respectively, and overall $mAP_{50}$ and $mAR_{50}$ increased by 0.021 and 0.020 respectively. It demonstrates that MAGFF effectively highlights object features and suppresses background interference in complex environments, confirming the effectiveness of MAGFF for feature fusion.

3) LFAMM: As an innovative learnable frequency domain processing module, LFAMM is designed to adjust and optimize the frequency components of images to improve the model’s ability to process blurry images. By comparing whether or not the LFAMM module is added, we observe that LFAMM significantly improves the performance of the model. Compared to the baseline model, the addition of the LFAMM module resulted in increases of 0.029, 0.0046, 0.002, and 0.006 in $AP_{50}$ across four categories, with accuracy increases of 0.019, 0.030, and 0.017 for small, medium, and large objects respectively. Overall, $mAP_{50}$ and $mAR_{50}$ increased by 0.021 and 0.022, respectively. Under the condition of using BRAB simultaneously, the addition of the LFAMM module improves the overall $mAP_{50}$ and $mAR_{50}$ by 0.019 and 0.023, respectively. Under the condition of using BRAB and MAGFF simultaneously, adding the LFAMM module improved $mAP_{50}$ and $mAR_{50}$ by 0.010 and 0.011, respectively. These results demonstrate the role of the LFAMM module in blurred image object detection, especially when detecting images with severe detail loss.

We compare the models with added modules to the Baseline and CenterNet-res50(Using ResNet50 as a backbone) in terms of floating-point operations (FLOPs) and parameter count (Para.), as shown in Table~\ref{table4}. From the table, it is evident that the FLOPs of the Baseline model (without any additional modules) are 193.24G, with a total parameter count of 17.93MB, significantly lower than the 235.75G FLOPs and 34.44MB of parameters of CenterNet-res50. This indicates that the DREB-Net model can effectively reduce the number of parameters while maintaining lower computational complexity. As the BRAB, MAGFF, and LFAMM modules are incrementally introduced, we observe a slight increase in FLOPs, but the growth in parameter count remains within a reasonable range. After integrating all three modules (BRAB, MAGFF, and LFAMM), the FLOPs are rising to 206.90G, and the parameter count increases to 18.97MB (30.28MB with BRAB deep branch added, which is only used during training and removed during inference), further validating the contribution of these modules to enhancing the model’s capabilities while keeping the computational costs relatively low.

Furthermore, we also present the original images, the object detection heatmaps generated using the baseline method and our proposed DREB-Net, as shown in Figure~\ref{heatmap}. Through these heatmaps, the advantages of DREB-Net in terms of precision and localization capabilities in object detection become more apparent. Compared to the baseline method, the heatmaps of DREB-Net show greater focus and more precise object center localization, indicating that our method can more accurately identify and locate objects within the images. Especially in complex backgrounds or blurry conditions, DREB-Net demonstrates exceptional resistance to interference and superior visual performance.

In conclusion, the addition of the three modules each demonstrated significant performance improvements, with the combination of BRAB, MAGFF, and LFAMM achieving overall optimal performance. These results not only validate the effectiveness of each module but also emphasize their important role in enhancing blurred image object detection.

\subsection{Testing Results on Real Blurry Images}
To validate the effectiveness of our proposed DREB-Net in practical application scenarios, we selected multiple real blurry images from the collected drone images for testing. These images encompass a range of real-world scenarios from varying lighting conditions to motion blur and long-distance shooting. These challenging scenarios enable a comprehensive assessment of the algorithm's capability to handle different types of blurring.

In the experiments, we carefully selected several representative real blurry image video clips and conducted object detection on these images using the DREB-Net model, which is pretrained on the VisDrone-2019-DET dataset. Figure~\ref{real_blur} shows these images and their results after DREB-Net processing, and Table~\ref{table_real_blur} provides a comparison of object detection results with CenterNet. The results indicate that our model can effectively recognize and locate objects in the images under complex blurry conditions. This demonstrates its adaptability and robustness in dealing with real-world blurry images, highlighting the practicality and potential of DREB-Net in a wide range of application scenarios.

\begin{table*}[htbp!]
	\centering
	\caption{Table for mAP and mAR testing of car on real blurry image video clips.}
	\label{table_real_blur}
	\setlength{\tabcolsep}{5pt}  
	\begin{tabular}{cccccccccc}
		\hline
		\multirow{2}{*}{{Scene}}  & \multicolumn{3}{c}{$mAP_{50}$} & \multicolumn{3}{c}{$mAR_{50}$}   & \multicolumn{3}{c}{FPS}  \\
		& {DREB-Net} & {DREB-Net\_tiny} & {CenterNet} & {DREB-Net}  & {DREB-Net\_tiny} & {CenterNet} & {DREB-Net}  & {DREB-Net\_tiny} & {CenterNet}  \\ \hline
		{Scene 1} & 0.684  & 0.663 & 0.593 & 0.513 & 5.509 & 0.486 & 11.63  & 31.93 & 26.88 \\ \hline
		{Scene 2} & 0.542 & 0.528 & 0.434 & 0.436 & 0.422 & 0.397 & 11.67  & 32.06 & 26.93 \\ \hline
		{Scene 3} & 0.562 & 0.554 & 0.443 & 0.447 & 0.438 & 0.413 & 11.58  & 31.91 & 26.90 \\ \hline
		{Scene 4} & 0.588 & 0.569 & 0.509 & 0.453 & 0.444 & 0.420 & 11.71  & 31.99 & 26.96 \\ \hline
		{Scene 5} & 0.573 & 0.562 & 0.485 & 0.436 & 0.424 & 0.399 & 11.66  & 32.11 & 26.99 \\ \hline
		{Scene 6} & 0.623 & 0.611 & 0.557 & 0.488 & 0.472 & 0.436 & 11.63  & 31.96 & 26.86 \\ \hline
	\end{tabular}
\end{table*}

\section{Conclusions}
\label{sec:conclusion}
In this paper, we introduced a novel object detection algorithm, DREB-Net, specifically designed and optimized for blurry images captured by unmanned aerial vehicles. The incorporation of the Multi-level Attention-Guided Feature Fusion(MAGFF) module, the Learnable Frequency domain Amplitude Modulation Module (LFAMM), and the Blurry image Restoration Auxiliary Branch (BRAB) significantly enhances the performance of object detection in complex blurry conditions. In our experiments, we evaluated the effectiveness of DREB-Net using two datasets, VisDrone-2019-DET and UAVDT. The results demonstrate that our method significantly improves both accuracy and recall rates compared to existing excellent algorithms, particularly in detecting small-sized objects. Additionally, comparative experiments and ablation studies further validated the effectiveness and necessity of the MAGFF module, the LFAMM module, and the BRAB.

Overall, our research shows that DREB-Net not only excels in standard object detection tasks but is also particularly effective in processing blurry images under complex conditions. Through detailed ablation studies, we confirmed the contribution of each module to enhancing overall system performance, especially in dynamic blurry environments. Future research could further explore the potential applications of these modules in other types of blurring and a broader range of image processing tasks. In conclusion, by precisely integrating multiple technological innovations, DREB-Net significantly improves the capability of object detection in drone images, which is crucial for applications such as drone video surveillance, traffic monitoring, and other scenarios requiring rapid and accurate object recognition. Further research and optimization of DREB-Net and its core modules will provide strong support for the development of UAV image processing technologies in the future.

\bibliographystyle{unsrt}
\bibliography{mybibfile}

\clearpage
\end{document}